\documentclass[journal]{IEEEtran}
\PassOptionsToPackage{nocompress}{cite}
\usepackage{amsmath,amsfonts}
\usepackage{bm}
\usepackage{algorithmic}
\usepackage{algorithm}
\usepackage{array}
\usepackage[caption=false,font=normalsize,labelfont=sf,textfont=sf]{subfig}
\usepackage{url}
\usepackage{verbatim}
\usepackage{graphicx}
\usepackage{cite}
\usepackage{pifont}  
\usepackage{xcolor}  
\usepackage{multirow}
\usepackage{tabularx,booktabs}
\usepackage[colorlinks=true, linkcolor=blue, citecolor=blue, urlcolor=black]{hyperref}
\begin{document}

\title{Spectral-Spatial Synergistic Guided Network \\ for Hyperspectral Salient Object Detection}

\author{Yanyan Peng, Tingfa Xu*, Yao Xiao, Peifu Liu, Shuyan Bai, Fengxiang Xu, and Jianan Li*
\thanks{
\textit{(Corresponding authors: Tingfa Xu; Jianan Li.)}}

\thanks{Yanyan Peng and Yao Xiao are with Chongqing Innovation Center, Beijing Institute of Technology, Chongqing 401120, China, and also with the School of Optics and Photonics, Beijing Institute of Technology, Beijing 100081, China (e-mail: 3220230741@bit.edu.cn; nuliweixiao0717@163.com).}

\thanks{Peifu Liu, Shuyan Bai, and Fengxiang Xu are with the School of Optics and Photonics, Beijing Institute of Technology, Beijing 100081, China (e-mail: 3120245389@bit.edu.cn; 3120230552@bit.edu.cn; elibest@163.com).}

\thanks{Tingfa Xu and Jianan Li are with Chongqing Innovation Center, Beijing Institute of Technology, Chongqing 401120, China, and with the School of Optics and Photonics, Beijing Institute of Technology, Beijing 100081, China, and also with the Key Laboratory of Photoelectronic Imaging Technology and System, Ministry of Education of China, Beijing 100081, China (e-mail: ciom\_xtf1@bit.edu.cn; lijianan@bit.edu.cn).}}



\maketitle

\begin{abstract}
Hyperspectral salient object detection aims to identify visually salient regions from hyperspectral images. Existing methods often fail because they fundamentally misunderstand the data, confusing incidental spectral variations caused by external factors such as illumination with essential spectral differences caused by the intrinsic material properties of the object. This leads to fragile representations and noisy predictions. To this end, we propose a lightweight and efficient Spectral-Spatial Synergistic Guided Network (S3GNet), with structure perception as the core, to build a closed-loop information flow around spectrum robust modeling, cross-stream co-perception and multi-scale refinement decoding. S3GNet introduces a parameter-free Spectral Structure-Aware Module that leverages spectral derivatives and regional hierarchical modeling to extract intrinsic features of robustness against illumination variations. Our Stream-Aware Attention Module achieves effective spectral-spatial collaboration through inter-stream global interaction and intra-stream spatial guidance. Furthermore, a Progressive Gated Refinement Decoder ensures precise object boundaries and detail recovery by optimally integrating multi-scale features. Experimental results show that S3GNet achieves superior performance in both computational efficiency and detection accuracy compared to existing methods. Code is available at https://github.com/pppyy0799/S3GNet.
\end{abstract}


\section{Introduction}
Salient object detection (SOD) aims to identify the most visually salient regions in an image~\cite{CPD,zhou2022multispectral,zhou2023position}. Early methods\cite{zhou2023lsnet,wang2024alignment}, relying mainly on RGB features, struggled to handle challenging scenes. In contrast, hyperspectral salient object detection (HSOD) utilizes the rich spectral information in hyperspectral images (HSIs) to distinguish objects based on their intrinsic material properties~\cite{SMN,ch1}, greatly improving the detection accuracy in complex scenes, with a wide range of applications~\cite{L1,xy,tian2}.

\begin{figure}[t]
  \centering
  \includegraphics[width=\columnwidth]{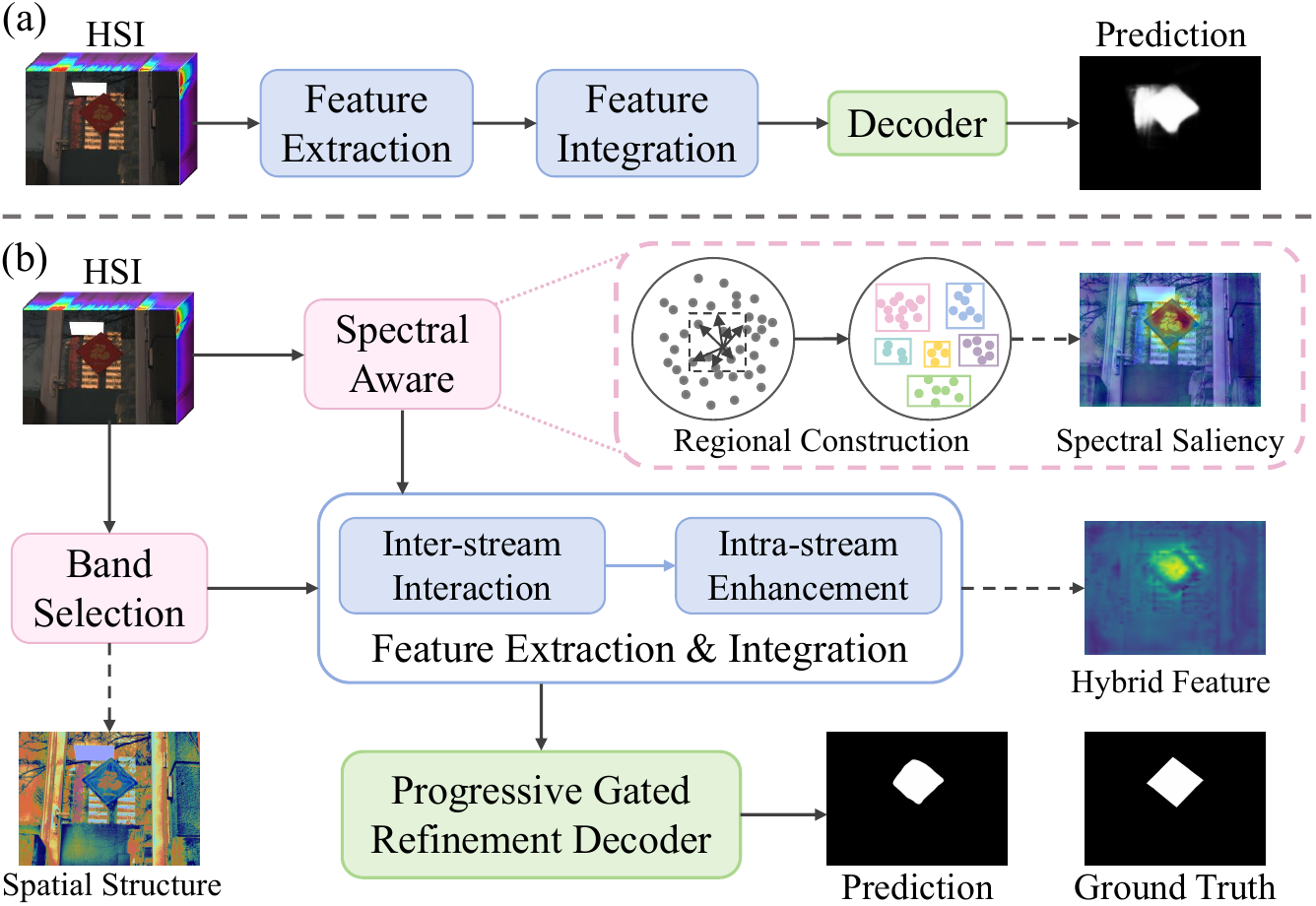} 
  \caption{Existing methods (a) are difficult to effectively distinguish between essential and incidental spectral differences, resulting in poor performance. Our method (b) extracts robust spectral representations and spatial structural features, and combines collaboration fusion and detail refinement strategies to achieve accurate and robust detection.}  
  \label{fig1_motiv}
\end{figure}

Despite this potential, current methods~\cite{SMN,L2} often fail to fully utilize HSI, as shown in Fig.~\ref{fig1_motiv}~(a). The core challenge lies in a fundamental ambiguity: a failure to effectively distinguish essential spectral differences, which arise from an object's intrinsic material properties, from incidental spectral differences, caused by external factors like illumination variance. This core challenge manifests in three critical limitations:
\textbf{(i) Fragile Feature Representation}: Models~\cite{ch,Qiu} relying on original spectral features exhibit severe vulnerability to illumination variations, easily mistaking incidental disturbances as intrinsic material differences, leading to unstable and inaccurate predictions.
\textbf{(ii) Ambiguous Cross-Stream Fusion}: Ineffective fusion strategies~\cite{SMN,Qin} fail to use spatial context to eliminate spectral ambiguities and ignore the intrinsic differences between the two streams, resulting in confusing representations and increased artefacts.
\textbf{(iii) Lossy Information Aggregation}: Naive aggregation techniques~\cite{8682522,Qin} tend to over-smooth features, treating subtle but critical essential spectral differences in small objects as incidental noise, significantly weakening the detection of small or structurally complex targets.

To address these limitations, we propose the Spectral-Spatial Synergistic Guided Network (S3GNet), an end-to-end framework integrating spectral awareness, cross-stream synergism, and refined decoding. Its dual-branch encoder extracts spatial features through band selection and robust spectral features via our Spectral Structure-Aware Module. These parallel features are then progressively fused by a Stream-Aware Attention Module and refined by a Progressive Gated Refinement Decoder to produce a precise saliency map, as shown in Fig.~\ref{fig1_motiv}~(b).

To overcome the sensitivity to illumination and enhance spectral discriminability, we introduce the Spectral Structure-Aware Module (SSAM). Instead of using original reflectance values, SSAM computes spectral derivatives to capture intrinsic spectral trends that are more indicative of material composition and less affected by lighting variations. This strategy effectively suppresses background noise. On this basis, it applies a parameter-free superpixel clustering strategy to aggregate these derivative features into regionally coherent representations, effectively preventing false negatives in homogeneous regions.

For effective cross-stream collaboration, we introduce Stream-Aware Attention Module (SAAM) as the core fusion unit. SAAM promotes synergy between the spectral and spatial streams through two specialized submodules. First, Weighted Correlation Attention (WCA) dynamically adjusts the contribution of features from both spectral and spatial streams by capturing the statistical correlation between the two streams, thus realizing the adaptive selection of each stream during the fusion process. Concurrently, Coupled Enhancement Attention (CEA) performs intra-stream spatial refinement, leveraging contextual cues from one stream to guide and enhance target features in the other, thus ensuring stronger complementarity and discriminative power of the fused features.

To enhance edge localization and detail recovery, we propose the Progressive Gated Refinement Decoder (PGRD). This decoder is architected around a Gated Refinement Module (GRM), which adaptively controls the information flow from shallow, high-resolution feature maps to deep, semantic ones. GRM is embedded within a progressive, bottom-up refinement strategy, allowing the network to incrementally restore intricate details and sharpen object boundaries at each stage of upsampling, yielding superior performance in localizing edges and detecting small targets.

In summary, S3GNet effectively solves the limitations of existing methods by introducing a branching feature extraction and fusion strategy of structure awareness and cross-stream alignment, combined with progressive refinement decoding, which significantly improves the model's detection performance in challenging scenarios such as light changes, complex backgrounds, and small targets.

Extensive experiments on benchmark datasets demonstrate that S3GNet sets a new state-of-the-art. Compared to the leading Hyper-HRNet~\cite{Qiu}, S3GNet increases $F_\beta$ by 10.8\%, all using only 33.0\% of the parameters and 48.0\% of the FLOPs. Furthermore, its strong performance on the RGB-T SOD task validates its cross-modal generalization capabilities.

Our contributions can be summarized as follows:
\begin{itemize}
  \item We propose a novel HSOD network, S3GNet, which significantly enhances detection performance by collaboratively leveraging spectral and spatial information. 
  \item We design a Spectral Structure-Aware Module to mine spectral derivatives and regional hierarchical cues to enhance the perception of intrinsic material properties.
  \item We introduce a Stream-Aware Attention Module, which achieves collaborative fusion through inter-stream global correlation and intra-stream spatial guidance.
  \item We construct a Progressive Gated Refinement Decoder that enhances the perception of small objects and edge regions by aggregating features at different levels.
\end{itemize}

\section{Related Work}
\subsection{Salient Object Detection}
Traditional SOD methods mainly rely on artificially constructed low-level visual features such as edges, colors, and spatial layout for salient region identification~\cite{jiang2013salient,rosin2009simple,achanta2009frequency}. For example, Rosin \textit{et al.}~\cite{rosin2009simple} performed pixel-level analysis through edge detection with threshold segmentation. Achanta \textit{et al.}~\cite{achanta2009frequency} proposed a frequency domain tuning model to enhance contrast sensitivity. Although these methods have some real-time performance in simple backgrounds, they often suffer from problems such as inaccurate target localization and blurred boundaries in complex backgrounds due to the lack of effective utilization of high-level semantic and structural information.

With the development of deep learning, CNN-based SOD methods have made significant progress~\cite{zhou2024frequency,tian1,tian2025tree}. Pang \textit{et al.}~\cite{pang2020multi} proposed a multi-scale self-interaction module to enhance context-awareness. Zhang \textit{et al.}~\cite{zhang2022r2net} used a residual pyramid structure to obtain rich semantics. In recent years, Sun \textit{et al.}~\cite{sun2023catnet} proposed a cascade aggregation transformer to alleviate the background interference problem to some extent. Unfortunately, these RGB methods are difficult to capture the spectral properties of HSI and have poor adaptability when migrating to HSOD, especially in complex materials and lighting environments with limited effects.

\subsection{Hyperspectral Salient Object Detection}
Although SOD has made significant progress in the field of RGB images, its research on HSI is still in the exploratory stage. Early HSOD methods were mostly inspired by Itti's model of visual attention\cite{Itti}. Liang \textit{et al.}\cite{6738493} used three-color combination to enhance spectral contrast. Moan \textit{et al.}\cite{Moan} extracted representative features by sub-band segmentation and PCA. However, these methods rely on downscaling and manual design, which are difficult to fully exploit the rich semantic information in HSI, resulting in poor performance.

\begin{figure*}[t]
  \setlength{\abovecaptionskip}{5pt}
  \centering
  \includegraphics[width=\textwidth]{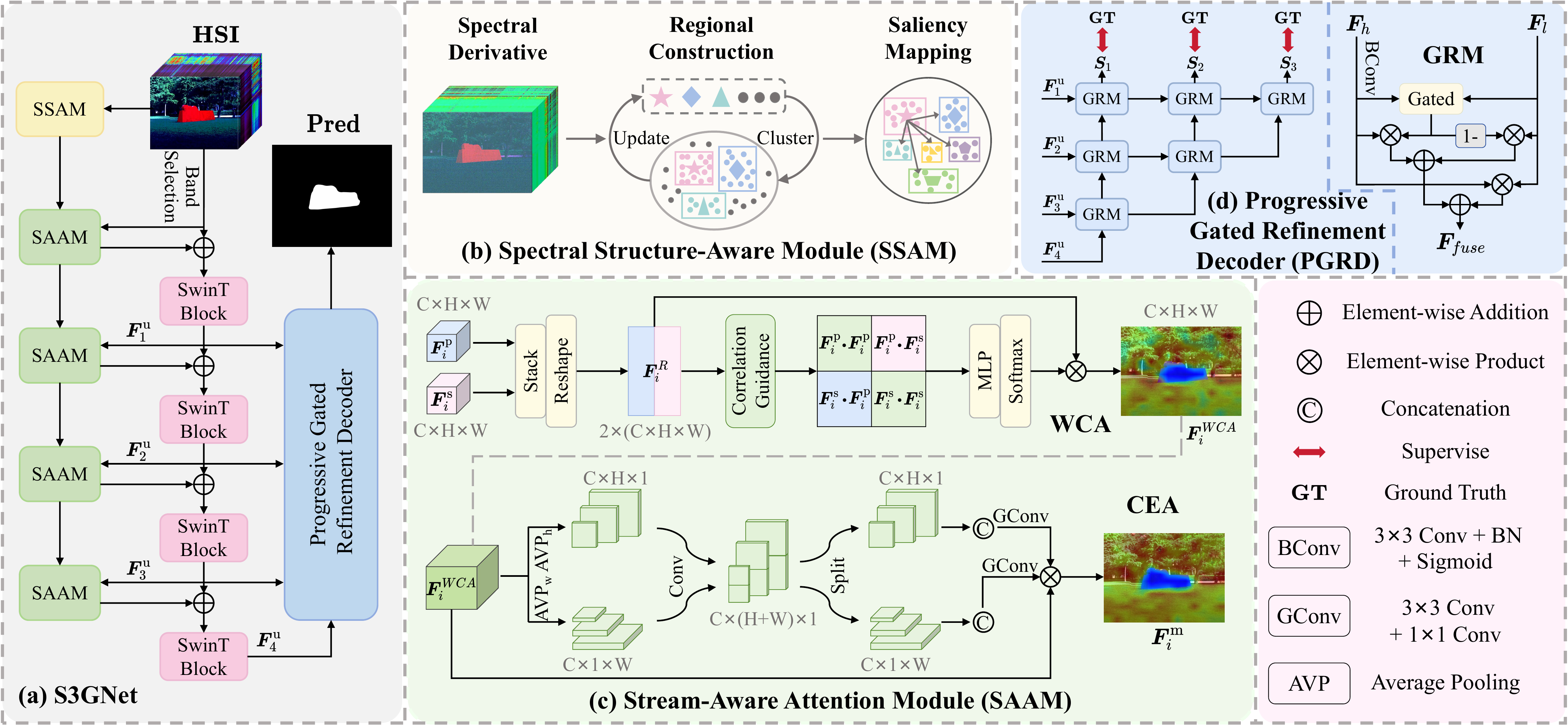} 
  \caption{(a) presents the overall architecture of the proposed S3GNet. It begins by leveraging (b) SSAM to extract spectral derivative information and hierarchical regional cues, while integrating band selection to capture spatial structural features. Subsequently, (c) SAAM performs deep fusion of multi-source representations by modeling global inter-stream correlations and intra-stream spatial guidance. Finally, (d) PGRD progressively aggregates multi-level features to generate high-quality saliency maps.}
  \label{fig2_all}
\end{figure*}

Recently, deep learning methods have been gradually introduced to the HSOD task to improve the detection performance. Huang \textit{et al.}\cite{ch} pioneered the construction of a dual-channel CNN to extract spatial and spectral features respectively to enhance the saliency expression. Liu \textit{et al.}\cite{SMN} combined the spectral saliency with the edge features to alleviate the problem of missing information. Qin \textit{et al.}\cite{Qin} introduced the knowledge distillation for dimensionality reduction. Qiu \textit{et al.}\cite{Qiu} used CNN and Transformer to extract and reconstruct spectral features. Although some progress has been made, challenges remain in capturing essential spectral differences, achieving cross-stream synergistic fusion, and restoring edge details. To this end, we propose a lightweight end-to-end network integrating spectral awareness, cross-stream synergy and refined decoding for more robust and accurate saliency detection.

\subsection{Attention Mechanism in Salient Object Detection}
In recent years, the introduction of attention mechanisms in SOD\cite{zhang2018progressive,liu2021visual,tian3} has significantly improved the robustness and localization accuracy of the model, including spatial-channel attention\cite{zhang2018progressive} and Transformer-based self-attention\cite{liu2021visual}. Subsequent studies have evolved around multi-scale fusion and cross-layer semantic enhancement\cite{yang2022bi}. 

In the field of HSOD, Liu \textit{et al.}~\cite{SMN} first introduced a mixed-frequency attention to improve spectral discrimination, and Qiu \textit{et al.}~\cite{Qiu} constructed a spectral attention reconstruction module to suppress redundancy. However, existing methods still exhibit limitations in capturing cross-stream complementary information. In contrast, we propose a novel attention mechanism to achieve more discriminative spectral-spatial feature fusion through inter-stream global interaction and intra-stream spatial guidance.

\section{Method}
Given an HSI $\bm{I} \in \mathbb{R}^{H \times W \times C}$, the objective of HSOD is to generate the corresponding saliency map $\bm{S} \in \mathbb{R}^{H \times W \times 1}$ using the mapping function $\boldsymbol{\Phi}(\cdot)$, highlighting objects or regions with saliency. This process can be expressed as:
\begin{equation}
\label{deqn_ex1}
 \bm{S} = \boldsymbol{\Phi}(\bm{I}).
\end{equation}

To implement the mapping function $\boldsymbol{\Phi}(\cdot)$, we propose a Spectral-Spatial Synergistic Guided Network (S3GNet). As shown in Fig.~\ref{fig2_all}, S3GNet consists of three main parts: the Spectral Structure-Aware Module (SSAM), the Stream-Aware Attention Module (SAAM), and the Progressive Gated Refinement Decoder (PGRD). 
Firstly, band selection and SSAM are used to extract the specialized spatial and spectral structure streams. Subsequently, SAAM intelligently fuses these streams and inputs them as prompts together with the spatial streams into the encoder to extract multi-scale features $\boldsymbol{F}^\text{u}=\left\{\boldsymbol{F}_{i}^\text{u}\right\}_{i=1}^{4}$. Notably, subsequent levels of SAAM are guided by low-level semantics to generate more complementary cross-stream representations. Ultimately, PGRD integrates the multi-scale features step by step to generate a clearly bounded and semantically consistent saliency map.

\subsection{Spectral Structure-Aware Module}
Since the original spectral features in HSI are easily affected by illumination variations, it is difficult to stably reflect the intrinsic material properties of objects. Therefore, we propose the Spectral Structure-Aware Module (SSAM), which utilizes spectral derivative features to replace the original reflectance and further combines the region-level aggregated contrast modeling, thereby enhancing the structural perception of salient targets. SSAM, as a plug-and-play standalone layer, is responsible for generating the spectral saliency a priori map as an input to S3GNet. It is shown in Fig.~\ref{fig2_all}~(b).

\noindent\textbf{Spectral Derivative Calculations.}
Derivatives of different orders exhibit different stability and noise properties\cite{derivative,wavelets}. The first-order derivative emphasizes the monotonic absorption trend and slope change in the reflectance curve, which is closely related to the inherent properties of the material. Although higher-order derivatives can capture curvature information, they are extremely sensitive to high-frequency noise. Therefore, we adopt the first-order spectral derivative to establish the spectral derivative feature map.

We denote the spectral features as $\boldsymbol{I}=\left\{\boldsymbol{v}_{i}\right\}_{i=1}^{C}$, where $\boldsymbol{v}_{i}$ is the reflectance matrix of the ${i}$-th band. The derivative of the ${i}$-th band can be defined as: 
\begin{equation}
\label{deqn_ex2}
\boldsymbol{g}_{i}= \frac{\boldsymbol{v}_{i+1}-\boldsymbol{v}_{i}}{\Delta \lambda },
\end{equation}
where $\Delta \lambda$ denotes the wavelength difference between two adjacent bands, which normalizes the derivative to a wavelength-invariant reflectance slope, thus improving the robustness of the cross-sensor. By calculating the difference in spectral derivatives between all bands, a spectral derivative map can be constructed: $\boldsymbol{G}=\left\{\boldsymbol{g}_{i}\right\}_{i=1}^{C-1}\in \mathbb{R}^{H \times W \times (C-1)}$.

\noindent\textbf{Regional Hierarchical Construction.}
Pixel-level saliency often lacks global consistency and is difficult to support the overall structure perception of the target. To improve structural consistency and accelerate processing, we introduce an improved CUDA-accelerated SLIC~\cite{SLIC} superpixel method to construct regional hierarchies based on the integration of spectral and spatial structures. The feature vector of each pixel is defined as: $\boldsymbol{f}_{p}=\left [ \boldsymbol{G}_{p},\frac{x_p}{m},\frac{y_p}{m} \right ] $, where $\left ( x_p,y_p \right )$ denotes the positional coordinates of the pixel, and m is a spatial weighting factor controlling the tightness of the clustering. In each iteration, we assign pixels to the nearest cluster centers based on the weighted Euclidean distance:
\begin{equation}
\label{deqn_ex3}
d(p,i) = \left \| \boldsymbol{G}_{p}- \boldsymbol{G}_{i} \right \|_2 +  \frac{1}{m} \left \| (x_p,y_p)-(x_i,y_i) \right \|_2,
\end{equation}
where $\boldsymbol{G}_{i}$ denotes the center feature vector. Then each superpixel center is updated and finally the image is divided into ${N}$ superpixel regions $\left \{{s_1,s_2,...s_N} \right \}$. For each superpixel ${s_i}$, its region is represented as the average of all internal pixel spectral derivatives:
\begin{equation}
\label{deqn_ex4}
\mu_i = \frac{1}{\left | s_i \right | } \sum_{\substack{p \in s_i }} \boldsymbol{G}_{p},
\end{equation}
where $\left | s_i \right |$ denotes the number of pixels in that region. 

\noindent\textbf{Spectral Saliency Mapping.}
To evaluate the spectral saliency difference between superpixel regions, the Euclidean distance between region average features is used as the contrast measure:
\begin{equation}
\label{deqn_ex5}
D_{ij} = \left \| \mu_i-\mu_j \right \|_2,
\end{equation}
Furthermore, the saliency score of the superpixel is defined as the weighted average of the contrast between the area and all other areas:
\begin{equation}
\label{deqn_ex6}
S_i=\sum_{\substack{j \neq i }}\omega_jD_{ij},\; \omega_j=\frac{\left | s_j \right | }{\sum_{k \neq i}\left | s_k \right | } 
\end{equation}
Finally, the saliency score of each superpixel $S_i$ is mapped back to the pixels it contains to obtain the pixel-level spectral saliency map $\bm{S_m} \in \mathbb{R}^{H \times W \times 1}$.

\subsection{Band Selection Strategy}
Due to the high spectral similarity of HSI between neighboring bands, significant redundant information is usually generated. To this end, we introduce an optimal neighborhood reconstruction (ONR)~\cite{ONR} method to retain its most discriminative spatial cues. The ONR screen out the key bands with the highest information content by evaluating the ability of candidate bands to reconstruct the entire spectral response. On this basis, we further employ a color mapping strategy to convert these representative bands into pseudo-color representations, resulting in spatial feature maps with clear edge and texture details.

\subsection{Stream-Aware Attention Module}
We design a Stream-Aware Attention Module (SAAM) that combines Weighted Correlation Attention and Coupled Enhancement Attention. Starting from the inter-stream global correlation and the intra-stream response respectively, it flexibly regulates the complementary relationship between spatial and spectral features, thereby enhancing the accuracy and robustness of salient target perception.

\noindent\textbf{Weighted Correlation Attention.}
Weighted Correlation Attention (WCA) captures inter-stream statistical correlations by constructing correlation matrices, and combines with multi-layer perceptron (MLP) to adaptively regulate feature contributions, effectively mitigating heterogeneous redundancies and differences, and enhancing feature complementarity and fusion accuracy. Specifically, for the input features $ \left\{\boldsymbol{F}_{i}^\text{p},\boldsymbol{F}_{i}^\text{s}\right\}_{i=1}^{4}$:
\begin{gather}
\boldsymbol{F}_{1}^\text{p} = \bm{I_p}, \;
\boldsymbol{F}_{i+1}^\text{p} = \boldsymbol{F}_{i}^\text{u},\quad i = 1,2,3,\\
\boldsymbol{F}_{1}^\text{s} = \bm{S_m}, \;
\boldsymbol{F}_{i+1}^\text{s} = \boldsymbol{F}_{i}^\text{m}, \quad i = 1,2,3,
\end{gather}
where $\bm{I_p}$ and $\bm{S_m}$ respectively represent the pseudo-color image feature and the spectral saliency image feature, $ \boldsymbol{F}_{i}^\text{u}$ represents the output feature from the ${i}$-th encoder layer, and $\boldsymbol{F}_{i}^\text{m}$ represents the output feature from the ${i}$-th SAAM block. We first stack and reshape $\boldsymbol{F}_{i}^\text{p}$ and $\boldsymbol{F}_{i}^\text{s}$ into a compact two-stream matrix $\boldsymbol{F}_{i}^{R}$:
\begin{equation}
\label{deqn_ex9}
\boldsymbol{F}_{i}^{R} = \mathrm{Reshape} (\mathrm{Stack(\boldsymbol{F}_{i}^\text{p},\boldsymbol{F}_{i}^\text{s})})\in \mathbb{R}^{2 \times D},
\end{equation}
where $D=C\times H\times W$ denotes the flattened spatial dimension.

Correlation Guidance: To eliminate the influence of scale differences across bands and streams, we perform feature-wise normalization on $\boldsymbol{F}_{i}^{R}$:
\begin{equation}
\label{deqn_ex10}
\hat{\boldsymbol{F}}_{i}^{R}=\frac{\boldsymbol{F}_{i}^{R}-\mu_i}{\sqrt{\sigma_i^2+\varepsilon } },
\end{equation}
where $\mu_i$ and $\sigma_i^2$ denote the mean and variance computed across the feature dimension $D$, and $\varepsilon$ is a small constant for numerical stability. Next, we measure the global statistical relationship between the two streams using a correlation matrix $\boldsymbol{R}_{i}$:
\begin{equation}
\label{deqn_ex11}
\boldsymbol{R}_{i}=\hat{\boldsymbol{F}}_{i}^{R}\cdot (\hat{\boldsymbol{F}}_{i}^{R})^\mathrm{T}\in \mathbb{R}^{2 \times 2},
\end{equation}
the diagonal entries of $\boldsymbol{R}_{i}$ quantify the internal consistency within each stream, whereas the off-diagonal entries reflect the degree of linear dependence between spectral and spatial features.

Feature Aggregation: To dynamically regulate the contribution of each stream, we use a lightweight MLP to map $\boldsymbol{R}_{i}$ nonlinearly and apply a softmax layer to obtain the fusion weights $\boldsymbol{\omega}_i= (\omega_i^\text{p},\omega_i^\text{s})$. The final fused feature is as follows:
\begin{equation}
\label{deqn_ex12}
\boldsymbol{F}_{i}^{WCA}=\omega_i^\text{p}\cdot\boldsymbol{F}_{i}^\text{p}+\omega_i^\text{s}\cdot\boldsymbol{F}_{i}^\text{s}.
\end{equation}

WCA can dynamically adjust the feature fusion strategy based on the global statistical relationship between streams, effectively suppressing redundant noise while retaining discriminative information.

\noindent\textbf{Coupled Enhancement Attention.}
To further enhance the response ability of the fused features to multi-scale salient objects in complex scenes, we input the output feature of WCA into Coupled Enhancement Attention (CEA). CEA combines Coordinate Attention~\cite{Coordinate} with cross-dimensional direction guidance mechanism to extract multi-scale spatial context and direction dependency. Specifically, we first downsample $\boldsymbol{F}_{i}^{WCA}$ to generate contextual features at different scales:
\begin{equation}
\label{deqn_ex13}
\boldsymbol{F}_{i1},\boldsymbol{F}_{i2},\boldsymbol{F}_{i3}=\mathrm{Down_n} (\boldsymbol{F}_{i}^{WCA}),\quad \mathrm{n}=1,2,4, 
\end{equation}
where $\mathrm{Down_n} (x)$ indicates performing $\mathrm{n}$ downsampling operations on $x$. At each scale, global average pooling is performed along the horizontal and vertical directions to extract orientation-aware features, respectively. We then concatenate these features and compress the channel dimensions. Next, the compressed features are reduced to horizontal and vertical attentional feature maps:
\begin{equation}
\label{deqn_ex14}
\boldsymbol{a}_{h/w}^j=\mathcal{T}_{\mathrm{h/w} }  (\mathcal{C} (\mathrm{AVP_h}(\boldsymbol{F}_{ij}),\mathrm{AVP_w}(\boldsymbol{F}_{ij}))),\quad j=1,2,3, 
\end{equation}
where $\mathrm{AVP_x}(\cdot )$ denotes average pooling along the x direction, $\mathcal{C}(x,y)$ denotes concatenate $x$ and $y$, and performs convolution, batch normalization, and activation function operation. $\mathcal{T}_{\mathrm{x}}(\cdot )$ means split in the x-direction. We connect the features in each direction. To further enhance the information interaction and feature alignment between directions, we introduce group convolution for coupling processing and generate normalized direction attention weights:
\begin{equation}
\label{deqn_ex15}
\boldsymbol{\alpha}_{h/w} =\delta(\mathrm{GConv} (\mathrm{Cat} (\left \{ \boldsymbol{a}_{h/w}^j \right\}))),
\end{equation}
where $\mathrm{Cat}(\cdot )$ denotes concatenation, $\mathrm{GConv}(\cdot )$ includes a $3\times3$ and a $1\times1$ convolution, $\delta$ represents the Sigmoid activation function. Finally, multiply them by the input feature $\boldsymbol{F}_{i}^{WCA}$ to obtain the final enhanced feature $\boldsymbol{F}_{i}^\text{m}$:
\begin{equation}
\label{deqn_ex16}
\boldsymbol{F}_{i}^\text{m}=\boldsymbol{F}_{i}^{WCA}\times \boldsymbol{\alpha}_{h}\times \boldsymbol{\alpha}_{w}.
\end{equation}

CEA effectively enhances the network's expressive ability in complex backgrounds and multi-scale targets by introducing long-range dependencies in spatial directions.

\subsection{Progressive Gated Refinement Decoder}
To improve the boundary clarity and detail integrity of the objects, We propose a Progressive Gated Refinement Decoder (PGRD). It is built around a Gated Refinement Module (GRM) embedded in a progressive refinement strategy.

\noindent\textbf{Gated Refinement Module.}
GRM adaptively controls the fusion strength of shallow spectral texture and deep semantic response by introducing a dynamic gating mechanism, thereby suppressing redundant information interference. As shown in Fig.~\ref{fig2_all}~(d), the high-level feature $\boldsymbol{F}_h$ is first upsampled to match the spatial resolution of the low-level features $\boldsymbol{F}_l$, and the channel dimension is adjusted by convolution to obtain the intermediate feature representation $\boldsymbol{F}_{h'}$:
\begin{equation}
\label{deqn_ex17}
\boldsymbol{F}_{h'} = \mathrm{BConv} (\mathrm{Up} (\boldsymbol{F}_h)),
\end{equation}
where $\mathrm{BConv}(\cdot)$ denotes $3\times3$ convolution, batch normalization and activation function. Then, we introduce a gating mechanism, which adaptively adjusts the fusion ratio of deep and shallow features by Sigmoid function, to obtain aggregated features with both global semantics and local details. Finally, in order to strengthen the salient objects and suppress the redundant information, the multiplicative interaction term is introduced to realize bidirectional enhancement. The whole process is as follows:
\begin{gather}
\boldsymbol{\alpha}_g=\delta (\boldsymbol{F}_{h'} + \boldsymbol{F}_l),\\
\boldsymbol{F}_{fuse}=\boldsymbol{\alpha}_g\times \boldsymbol{F}_{h'}+ (1-\boldsymbol{\alpha}_g)\times \boldsymbol{F}_l+ \boldsymbol{F}_{h'}\times  \boldsymbol{F}_l.
\end{gather}

\noindent\textbf{Progressive Refinement Strategy.}
To further mitigate the feature loss problem during upsampling, we introduce a bottom-up progressive refinement strategy. By gradually fusing the saliency maps of the middle layers, we guide the high-level features to recover spatial details and realise salient target reconstruction from coarse to fine. Finally, high-quality prediction maps are generated.

\begin{table*}[t]
\caption{Quantitative Results on HSOD-BIT-V2 Dataset}
\label{tab:table1}
\centering
\begin{tabular*}{\textwidth}{@{\extracolsep{\fill}}%
@{\hspace{12pt}}>{\raggedright\arraybackslash}p{2.2cm}  
|>{\centering\arraybackslash}p{1cm}  
>{\centering\arraybackslash}p{1cm}  
>{\centering\arraybackslash}p{1cm}  
>{\centering\arraybackslash}p{1cm}  
>{\centering\arraybackslash}p{1cm}  
>{\centering\arraybackslash}p{1cm}@{\hspace{16pt}}  
|>{\centering\arraybackslash}p{1.2cm}   
>{\centering\arraybackslash}p{1.2cm}    
>{\centering\arraybackslash}p{1.1cm}@{\hspace{8pt}} 
}
\toprule
Method & $\mathrm{MAE} \downarrow$ & $\mathrm{PRE}\uparrow$ & $\mathrm{REC}\uparrow$ & $E_\xi\uparrow$ & $F_\beta\uparrow$ & $S_\alpha\uparrow$ & \#Params & FLOPs & FPS \\
\midrule
\multicolumn{10}{c}{\textit{RGB-based SOD Methods}} \\
\midrule
Itti~\cite{Itti} & 0.230 & 0.280 & 0.419 & 0.606 & 0.245 & 0.449 & - & - & - \\
BASNet~\cite{qin2019basnet} & 0.049 & 0.638 & 0.634 & 0.751 & 0.557 & 0.635 & 87.06 M & 127.56 G & 106.06 \\
U2Net~\cite{qin2020u2} & 0.046 & 0.649 & 0.597 & 0.749 & 0.507 & 0.582 & 44.01 M & 47.65 G & 130.95 \\
ICON~\cite{zhuge2022salient} & 0.051 & 0.662 & 0.574 & 0.782 & 0.532 & 0.524 & 33.09 M & 8.49 G & 118.72 \\
MDSAM~\cite{gao2024} & 0.047 & 0.641 & 0.587 & 0.768 & 0.544 & 0.620 & 100.21 M & 66.23 G & 69.25 \\
SelfReformer~\cite{SelfReformer} & 0.031 & 0.657 & 0.628 & 0.802 & 0.635 & 0.779 & 90.70 M & 12.82 G & 99.24 \\
ADMNet~\cite{ADMNet} & 0.040 & 0.625 & 0.634 & 0.798 & 0.586 & 0.736 & 0.84 M & 0.82 G & 166.08\\
SDNet-A~\cite{SDNet} & 0.042 & 0.620 & 0.577 & 0.756 & 0.602 & 0.731 & 0.82 M & 5.12 G & 276.07\\
TIGAN~\cite{TIGAN} & 0.026 & 0.706 & 0.676 & 0.834 & 0.671 & 0.783 & 87.32 M & 46.54 G & 125.03 \\
\midrule
\multicolumn{10}{c}{\textit{HSI-based HSOD Methods}} \\
\midrule
SAD~\cite{6738493} & 0.180 & 0.355 & 0.372 & 0.559 & 0.251 & 0.447 & - & - & -\\
SED~\cite{6738493} & 0.112 & 0.356 & 0.185 & 0.573 & 0.232 & 0.452 & - & - & -\\
SG~\cite{6738493} & 0.177 & 0.342 & 0.350 & 0.548 & 0.240 & 0.451 & - & - & -\\
SUDF~\cite{8682522} & 0.166 & 0.382 & 0.622 & 0.590 & 0.364 & 0.589 & 0.10 M & 82.90 G & 0.24\\
SMN~\cite{SMN} & 0.039 & 0.607 & 0.713 & 0.747 & 0.573 & 0.706 & 10.23 M & 14.76 G & 89.18 \\
DMSSN~\cite{Qin} & 0.072 & 0.635 & 0.602 & 0.625 & 0.550 & 0.673 & 1.76 M & 10.89 G & 139.80\\
Hyper-HRNet~\cite{Qiu} & 0.028 & 0.653 & 0.764 & 0.806 & 0.591 & 0.784 & 29.57 M & 18.96 G & 58.52\\
\midrule 
S3GNet (Ours) &\textbf{0.021} & \textbf{0.710} & \textbf{0.768} & \textbf{0.859} & \textbf{0.699} & \textbf{0.801} & 9.74 M & 9.12 G & 137.26 \\
\bottomrule
\end{tabular*}
\end{table*}

\begin{figure*}[t] 
  \centering
  \includegraphics[width=\textwidth]{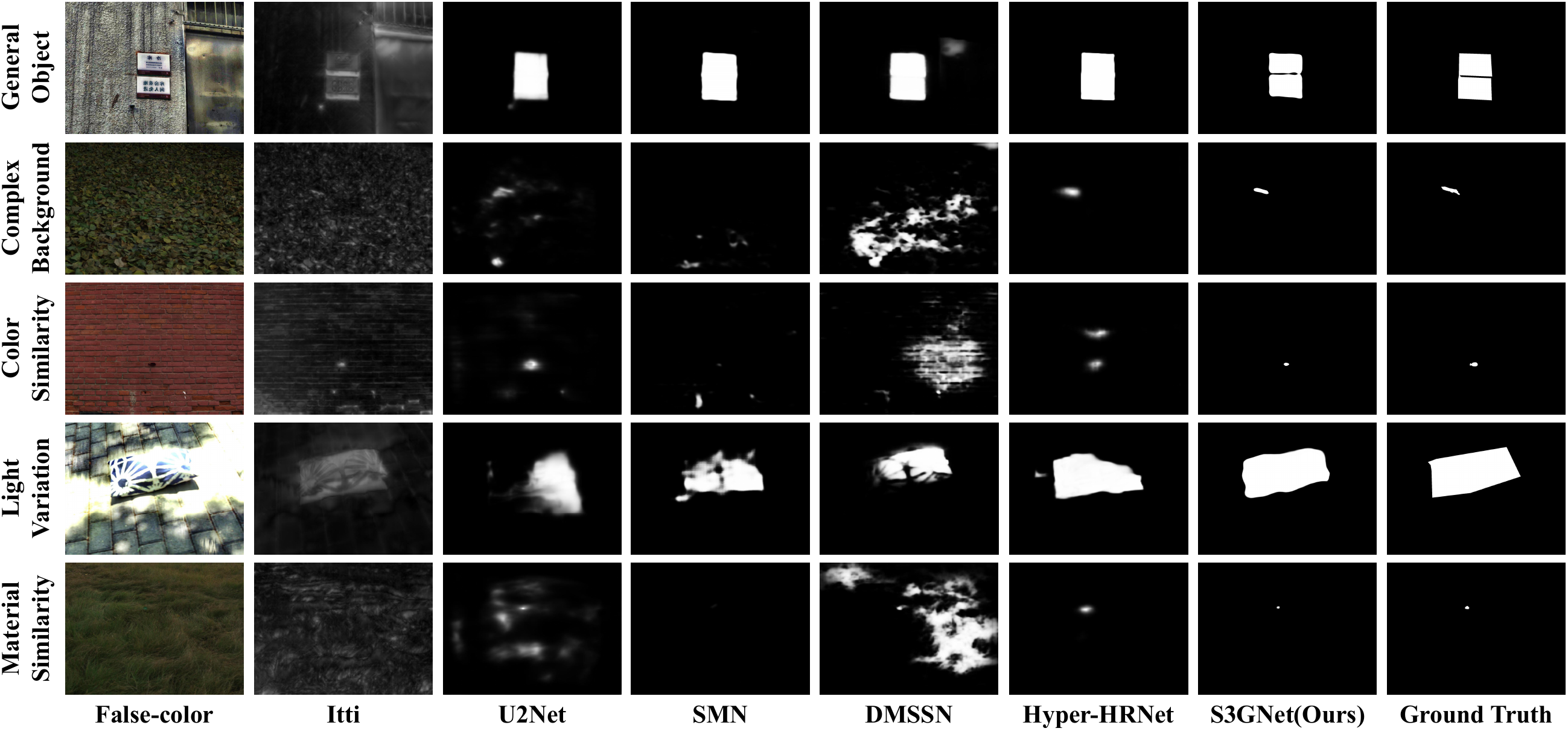} 
  \caption{Qualitative results on HSOD-BIT-V2 dataset. S3GNet has the best detection performance, especially in challenging scenarios.}
  \label{fig_hsodbit}
\end{figure*}

\subsection{Learning Objective}
We use a hybrid loss function consisting of a binary cross-entropy (BCE) loss function and an intersection over union (IoU) loss function to supervise the saliency prediction maps $\left \{ \boldsymbol{S}_i \right \}_{i=1}^3$ generated at each stage. The total loss is defined as:
\begin{equation}
\label{deqn_ex20}
\mathcal{L}=\sum_{i=1}^{3}\left [ \mathcal{L}_\text{BCE}^i(\boldsymbol{S}_i,\boldsymbol{G})+ \mathcal{L}_\text{IoU}^i(\boldsymbol{S}_i,\boldsymbol{G})\right ],
\end{equation}
where $\boldsymbol{G}$ denotes the ground-truth saliency map.

\begin{figure}[t] 
  \centering
  \includegraphics[width=\linewidth]{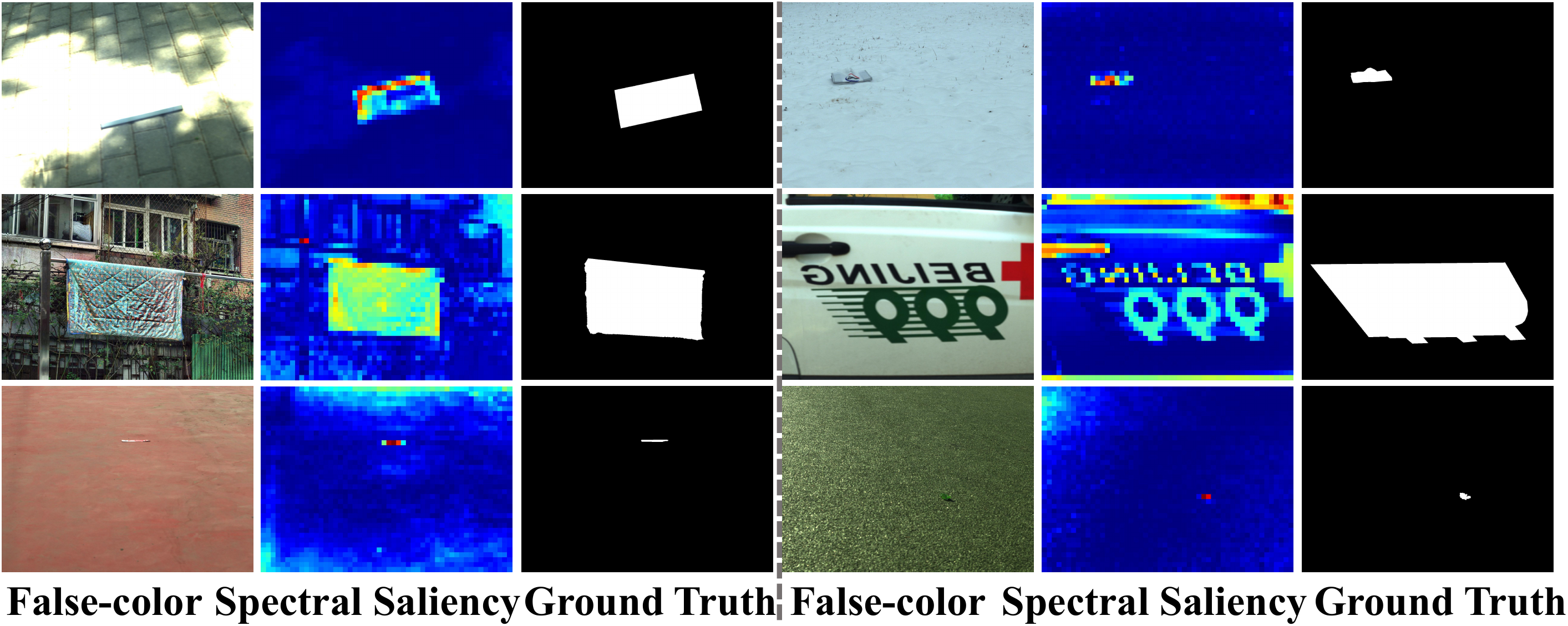} 
  \caption{Spectral saliency map visual of SSAM output. The module effectively highlights spectrally salient regions based on spectral derivative features, providing a high quality spectral prior for the model.}
  \label{fig_Spectral}
\end{figure}

\begin{figure}[t] 
  \centering
  \includegraphics[width=\linewidth]{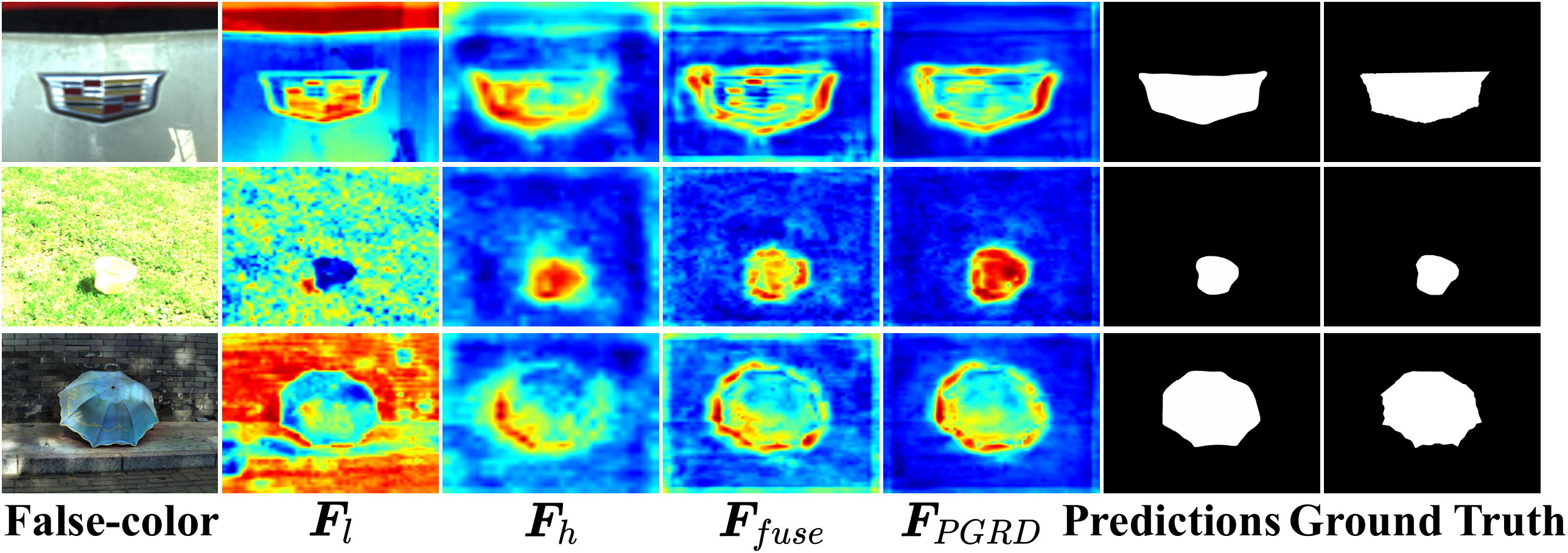} 
  \caption{Visualization of the feature refinement process in PGRD.}
  \label{fig_PGRD}
\end{figure}

\section{Experiments}
\subsection{Implementation Details}
To fully evaluate the saliency detection capabilities of S3GNet, we conduct experiments on HS-SOD~\cite{HSSOD} and HSOD-BIT-V2~\cite{Qiu} datasets and compare them with a variety of representative methods, including baseline models in the HSOD domain, traditional methods, and current open-source state-of-the-art methods. Our model is implemented based on the PyTorch framework and trained in an environment equipped with a single NVIDIA GeForce RTX 4090 GPU. All input HSIs are uniformly scaled to a spatial resolution of 224 × 224, and data enhancement strategies such as random horizontal flipping and random cropping are used to effectively mitigate the overfitting problem. The backbone network parameters of the model are initialized from the pre-trained weights on the ImageNet-1k dataset.

We measure the performance of our S3GNet using the following commonly used evaluation metrics~\cite{li2023delving,jiang2013salient}: mean absolute error (MAE), precision (PRE), recall (REC), E-measure ($E_\xi$), adaptive F-measure ($F_\beta$), and S-measure ($S_\alpha$). Together, these metrics constitute a comprehensive evaluation system of the performance of S3GNet, allowing us to make quantitative comparisons with existing HSOD methods and validate their effectiveness in HSOD.

\subsection{Results on HSOD-BIT-V2 Dataset}
\subsubsection{Data and Setups}
The HSOD-BIT-V2 dataset contains 500 high-quality HSIs with a spatial resolution of 1240 × 1680 pixels and a band spacing of 3 nm, covering the spectral range from 400 to 1000 nm. Among them, 406 HSIs are used for training and 94 for testing. During the training process, we use an Adam optimizer\cite{kingma2014adam} with an initial learning rate of $5 \times 10^{-5}$ and a batch size of 8, and train for a total of 200 epochs. To achieve more stable convergence, the learning rate is automatically decayed to 1/10 of the original rate after every 100 epochs.

\subsubsection{Quantitative Results}
As shown in the results of Table~\ref{tab:table1}, S3GNet achieves leading scores in all evaluation metrics on HSOD-BIT-V2 dataset. Specifically, it outperforms Hyper-HRNet in MAE, PRE, REC, $E_\xi$, $F_\beta$ and $S_\alpha$ by 0.7\%, 5.7\%, 0.4\%, 5.3\%, 10.8\% and 1.7\%, respectively, which demonstrates a significant performance advantage. Compared to the RGB-based method, which only relies on color cues, and the HSI-based method, which directly utilizes the original spectrum, S3GNet greatly improves detection accuracy and stability in complex scenes by distinguishing the essential spectral differences of objects from incidental spectral disturbances.

\subsubsection{Efficiency Analysis}
We evaluated the efficiency of S3GNet with several representative methods, as shown in Table~\ref{tab:table1}. RGB-based methods usually employ complex structures to enhance expressive power, resulting in larger parameter sizes. Although the lightweight models ADMNet and SDNet-A have small parameter sizes, their performance on HSI is significantly limited by the lack of spectral modeling. HSI-based methods focus more on efficiency and have relatively fewer parameters. Among them, SUDF has the fewest parameters but is computationally intensive with 82.9G FLOPs due to the introduction of flow learning, while DMSSN compresses the model through knowledge distillation, which is lightweight but not directly comparable. In contrast, S3GNet achieves 137.26 FPS with only 9.74M parameters and 9.12G FLOPs, which achieves the best accuracy while maintaining real-time performance, showing an excellent efficiency-performance balance.

\subsubsection{Qualitative Results}
It can be clearly seen from Fig.~\ref{fig_hsodbit} that the existing methods often encounter problems such as blurred boundaries and missed or false detections in complex scenarios. In rows 3 and 5, where colors and materials are highly similar or the target is small, the detection model struggles to effectively distinguish between the target and the background due to ignoring subtle essential spectral differences. In row 4, under strong lighting conditions, incidental illumination changes are mistakenly interpreted as saliency loss. In contrast, S3GNet effectively suppresses incidental disturbances by capturing essential spectral differences, while retaining key details and boundary information, achieving stable and accurate detection.

\subsubsection{Visualization of Spectral Saliency Maps}
As shown in Fig.~\ref{fig_Spectral}, the spectral feature maps generated by our proposed SSAM effectively highlight the spectrally salient object regions in the HSI. It not only improves the spectral contrast between foreground and background, but also preserves the key structural information, which provides strong feature support for subsequent salient region extraction.

\subsubsection{Visualization of PGRD}
Fig.~\ref{fig_PGRD} shows the visualized feature maps of PGRD. It can be found that the low-level feature map $\boldsymbol{F}_l$ retains the edge and texture information but is more noisy, and the high-level feature $\boldsymbol{F}_h$ focuses on the semantic region and spectral response but lacks details. The GRM fused feature $\boldsymbol{F}_{fuse}$ enhances semantic expressiveness and edge clarity while suppressing redundant interference. Combined with the progressive refinement strategy, the salient map output from PGRD achieves local structure recovery and global consistency, effectively improving the target localization accuracy.

\subsubsection{Visualization of Feature Distribution}
\begin{figure}[t] 
  \centering
  \includegraphics[width=\linewidth]{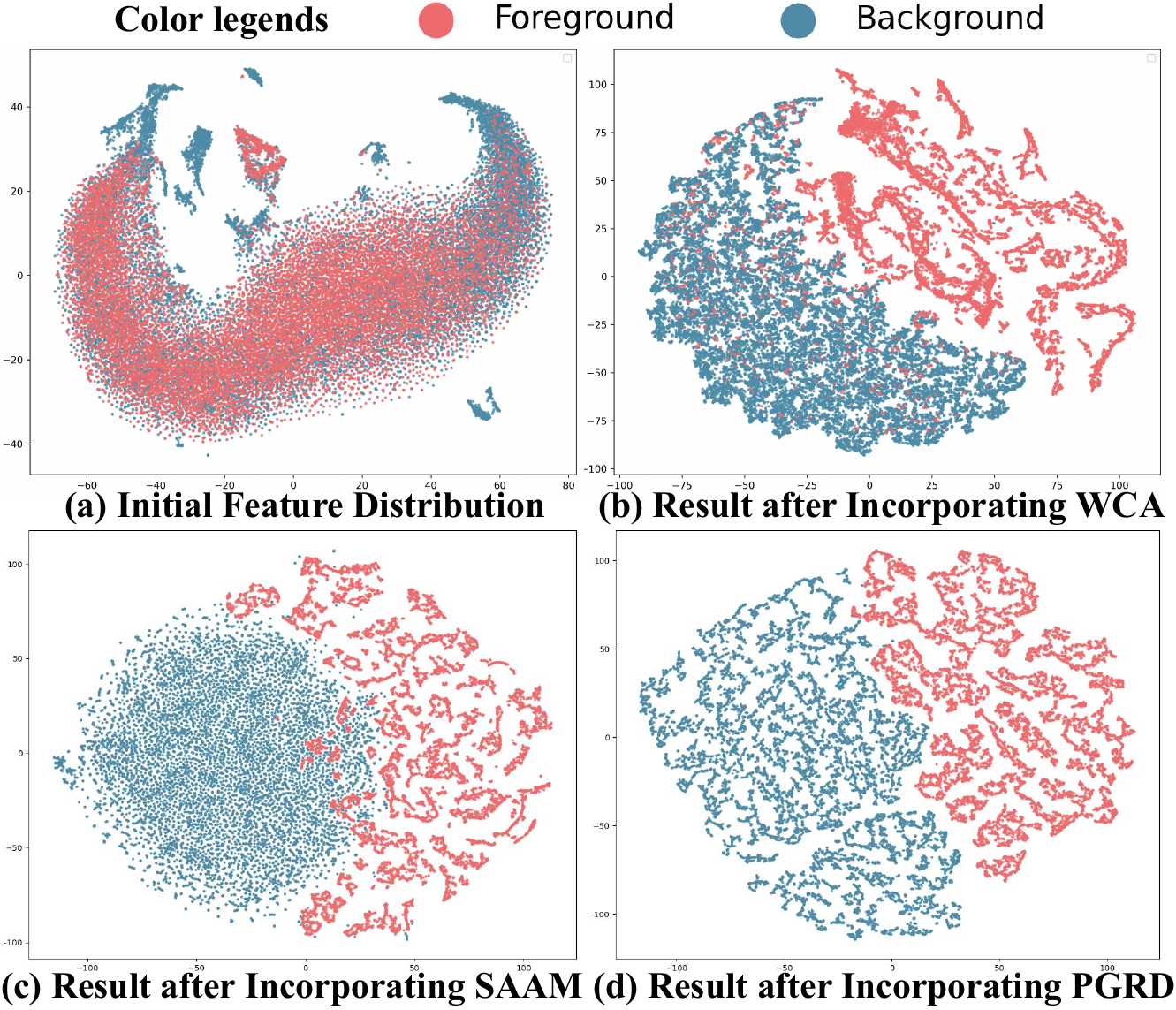} 
  \caption{Visualization of feature distribution via t-SNE. Red indicates salient objects, blue indicates background.}
  \label{fig_feafb}
\end{figure}

We use t-SNE to visualize the changes in feature distribution at each stage. As shown in Fig.~\ref{fig_feafb}, the initial features (a) are severely mixed with foreground and background, reflecting the confusion between essential spectral differences and incidental spectral disturbances. As each module acts sequentially, the features gradually aggregate in an orderly manner, ultimately achieving clear inter-class separation in (d).

\begin{table}[t]
\caption{Quantitative Results on HS-SOD Dataset\label{tab:table2}}
\centering
\begin{tabular*}{\linewidth}{@{\extracolsep{\fill}}@{\hspace{6pt}}>{\raggedright\arraybackslash}p{2.2cm}|ccccc}
\toprule
Method & $\mathrm{MAE} \downarrow$ & $\mathrm{PRE}\uparrow$ & $\mathrm{REC}\uparrow$ & $E_\xi\uparrow$ & $S_\alpha\uparrow$ \\
\midrule
Itti~\cite{Itti} & 0.245 & 0.367 & 0.332 & 0.570 & 0.545  \\
SAD~\cite{6738493} & 0.236 & 0.353 & 0.311 & 0.583 & 0.537  \\
SED~\cite{6738493} & 0.185 & 0.268 & 0.281 & 0.521 & 0.462  \\
SG~\cite{6738493} & 0.218 & 0.323 & 0.343 & 0.588 & 0.549  \\
SUDF~\cite{8682522} & 0.242 & 0.265 & 0.338 & 0.561 & 0.468  \\
SMN~\cite{SMN} & 0.069 & 0.624 & 0.767 & 0.810 & 0.757  \\
DMSSN~\cite{Qin} & 0.068 & 0.694 & 0.764 & 0.832 & 0.758  \\
Hyper-HRNet~\cite{Qiu} & 0.056 & 0.702 & 0.779 & 0.861 & 0.771 \\
\midrule 
S3GNet (Ours) &\textbf{0.051} & \textbf{0.715} & \textbf{0.787} & \textbf{0.882} & \textbf{0.773} \\
\bottomrule
\end{tabular*}
\end{table}

\newcommand{\tick}{\raisebox{0pt}[0.85em]{\color{green}\ding{51}}}
\newcommand{\cross}{\raisebox{0pt}[0.85em]{\color{red}\ding{55}}}

\begin{figure}[t]
  \centering
  \includegraphics[width=\linewidth]{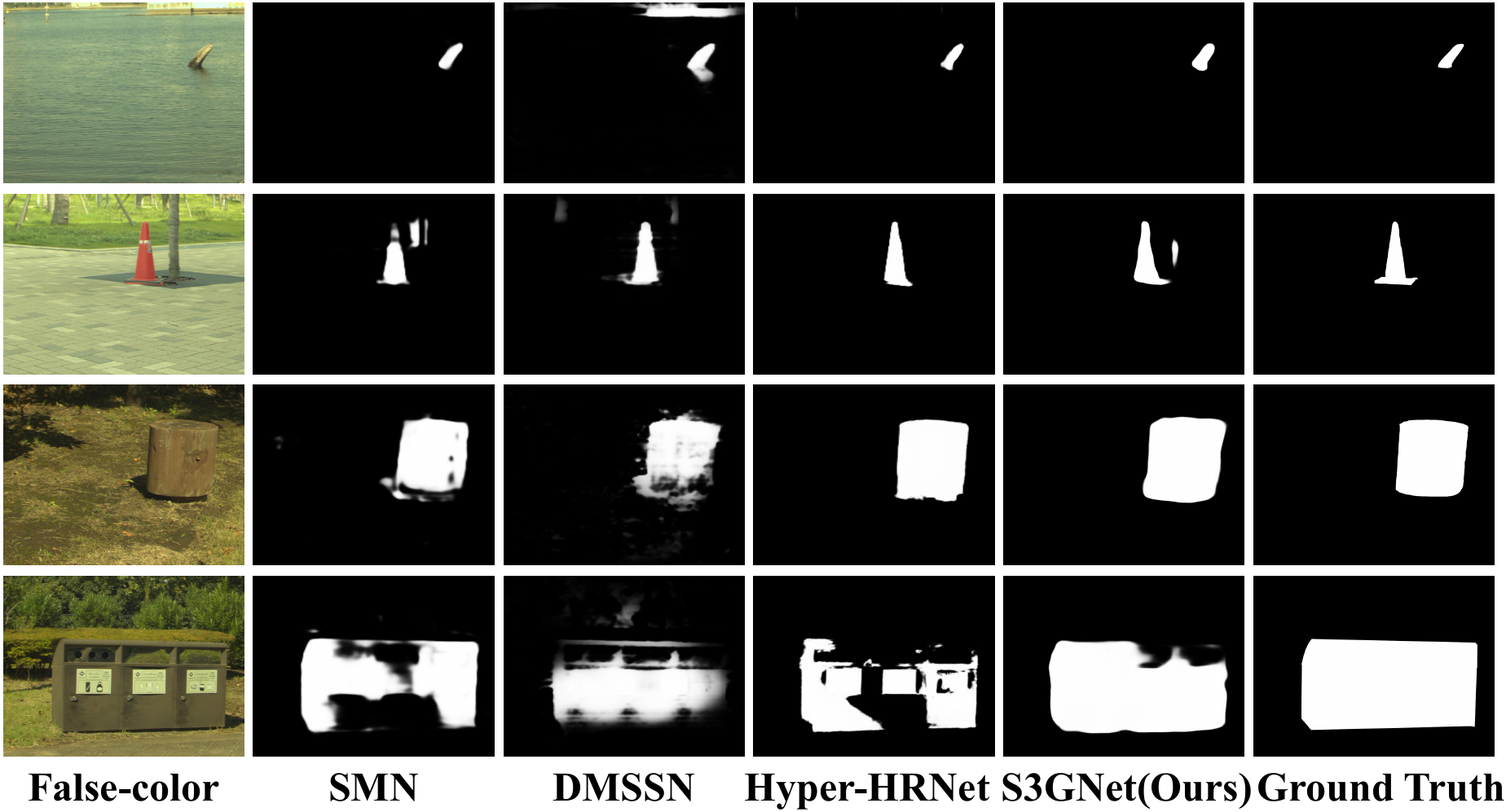} 
  \caption{Qualitative results on HS-SOD dataset. S3GNet can identify salient regions more efficiently.}
  \label{fig_hssod}
\end{figure}


\begin{table}[t]
\centering
    \begin{minipage}[t]{\linewidth}
        \centering
        \caption{Ablation Study of Key Components\label{tab:table03}}
        \setlength{\tabcolsep}{2pt} 
        \begin{tabular*}{\linewidth}{@{\extracolsep{\fill}}%
        >{\centering\arraybackslash}p{0.9cm}
        >{\centering\arraybackslash}p{0.9cm}
        >{\centering\arraybackslash}p{0.9cm}
        >{\centering\arraybackslash}p{0.9cm}
        |>{\centering\arraybackslash}p{1.1cm}
        >{\centering\arraybackslash}p{1.1cm}
        >{\centering\arraybackslash}p{1.1cm}
        >{\centering\arraybackslash}p{1.1cm}}
        
        \toprule
        Baseline & SSAM & SAAM & PGRD & $\mathrm{MAE}\downarrow$ & $E_\xi\uparrow$ & $F_\beta\uparrow$ & $S_\alpha\uparrow$ \\
        \midrule
        
        \tick & \cross & \cross & \cross
        & 0.076 & 0.734 & 0.523 & 0.680 \\
        
        \tick & \tick & \cross & \cross
        & 0.059 & 0.771 & 0.528 & 0.695 \\
        
        \tick & \tick & \tick & \cross
        & 0.045 & 0.780 & 0.533 & 0.701 \\
        
        \tick & \tick & \tick & \tick
        & \textbf{0.021} & \textbf{0.859} & \textbf{0.699} & \textbf{0.801} \\
        \bottomrule
    \end{tabular*}
    \end{minipage}

    \vspace{1em}

    \begin{minipage}[t]{\linewidth}
        \centering
        \caption{Ablation Study of Input Data\label{tab:table3}}
        \begin{tabular*}{\linewidth}{@{\extracolsep{\fill}}
        >{\centering\arraybackslash}p{2cm}
        @{}
        >{\centering\arraybackslash}p{1.3cm}
        |cccc}
        \toprule
        False-color & Spec. Sal. & $\mathrm{MAE} \downarrow$ & $E_\xi\uparrow$ & $F_\beta\uparrow$ & $S_\alpha\uparrow$ \\
        \midrule
        \makebox[2cm][c]{\color{green}\ding{51}} & \makebox[1.3cm][c]{\color{red}\ding{55}} & 0.026 & 0.816 & 0.634 & 0.764 \\
        \makebox[2cm][c]{\color{green}\ding{51}} & \makebox[1.3cm][c]{\color{green}\ding{51}} & \textbf{0.021} & \textbf{0.859} & \textbf{0.699} & \textbf{0.801} \\
        \bottomrule
        \end{tabular*}
    \end{minipage}
\end{table}

\begin{figure}[t] 
  \centering
  \includegraphics[width=\linewidth]{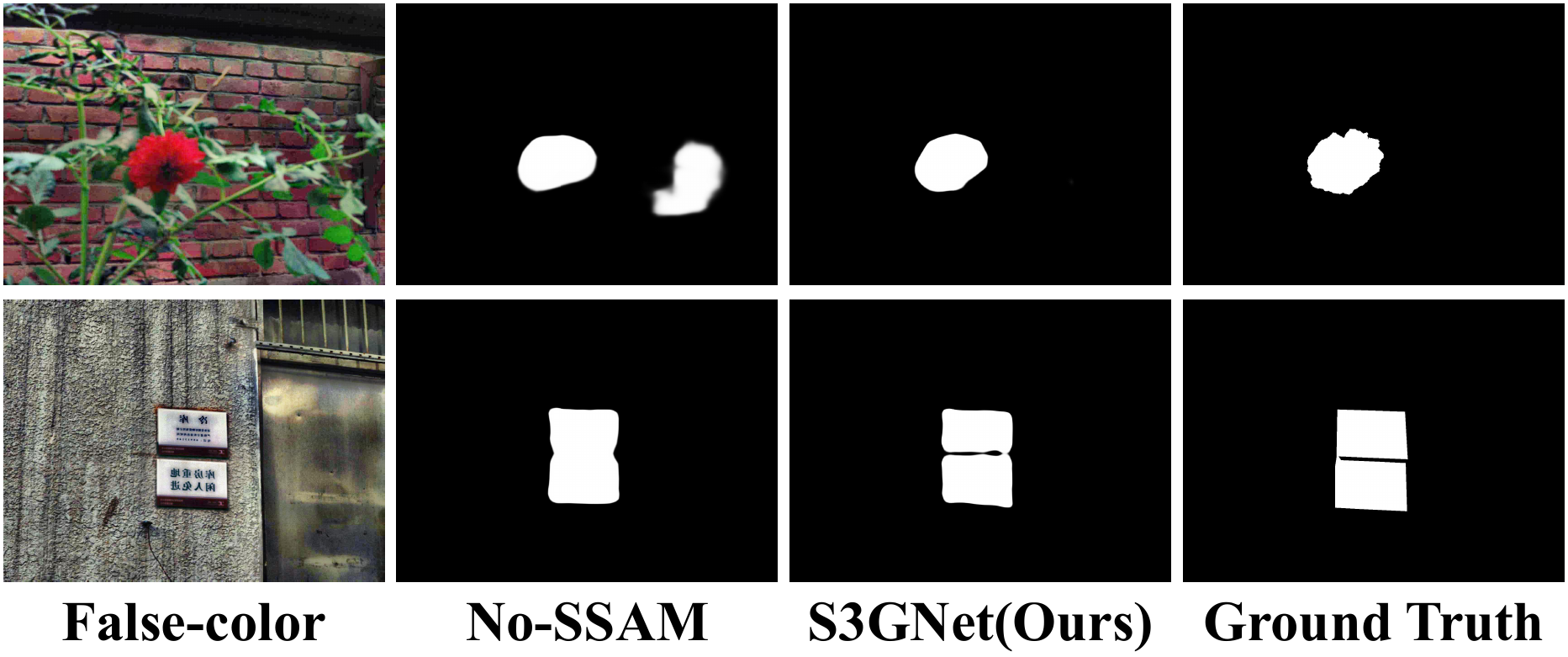} 
  \caption{Results with and without spectral features.}
  \label{fig_SSAM}
\end{figure}

\begin{table}[t]
    \centering
    \begin{minipage}[t]{\linewidth}
        \centering
        \caption{Impact of the Order of Spectral Derivatives\label{tab:table13}}
        \begin{tabular*}{\linewidth}{@{\extracolsep{\fill}}@{\hspace{5pt}}%
        >{\centering\arraybackslash}p{2.5cm}@{\hspace{6pt}}
        |cccc}
        \toprule
        Method & $\mathrm{MAE} \downarrow$ & $E_\xi\uparrow$ & $F_\beta\uparrow$ & $S_\alpha\uparrow$ \\
        \midrule
        Raw Spectrum & 0.025 & 0.850 & 0.675 & 0.788 \\
        First-order & \textbf{0.021} & 0.859 & \textbf{0.699} & \textbf{0.801} \\
        Second-order & 0.024 & 0.861 & 0.684 & 0.793 \\
        Mixed-order & 0.023 & \textbf{0.863} & 0.694 & 0.798 \\
        \bottomrule
        \end{tabular*}
    \end{minipage}

    \vspace{1em}

    \begin{minipage}[t]{\linewidth}
        \centering
        \caption{Ablation Study of Different Fusion Strategies\label{tab:table4}}
        \centering
        \begin{tabular*}{\linewidth}{@{\extracolsep{\fill}}@{\hspace{1pt}}%
        >{\centering\arraybackslash}p{1.3cm}
        @{}
        >{\centering\arraybackslash}p{1.3cm}@{\hspace{1pt}}
        |cccc}
        \toprule
        Method & Depth & $\mathrm{MAE} \downarrow$ & $E_\xi\uparrow$ & $F_\beta\uparrow$ & $S_\alpha\uparrow$ \\
        \midrule
        Baseline & - & 0.026 & 0.816 & 0.634 & 0.764 \\
        +Spectral & 1 & 0.026 & 0.796 & 0.606 & 0.756 \\
        +Spectral & 4 & 0.025 & 0.828 & 0.619 & 0.761 \\
        +SAAM & 1 & 0.023 & 0.843 & 0.674 & 0.783 \\
        +SAAM & 4 & \textbf{0.021} & \textbf{0.859} & \textbf{0.699} & \textbf{0.801}\\
        \bottomrule
        \end{tabular*}
    \end{minipage}
\end{table}

\subsection{Results on HS-SOD Dataset}
\subsubsection{Data and Setups}
The HS-SOD dataset contains 60 HSIs with a spatial resolution of 768 × 1024 pixels and a spectral sampling interval of 5 nm, covering the spectral range from 380 to 780 nm. According to the standard division strategy, 48 of the images are used for training and the remaining 12 are used for testing. We train a total of 100 epochs using the Adam optimizer with an initial learning rate of 5e-5 and a batch size of 4.

\subsubsection{Quantitative Results}
The quantitative comparison results on HS-SOD dataset are shown in Table~\ref{tab:table2}. Our proposed S3GNet outperforms existing RGB-based and HSI-based methods in all evaluation metrics, demonstrating superior detection capabilities. Specifically, S3GNet obtains 0.051, 0.715, 0.787, 0.882, and 0.773 in MAE, PRE, REC, $E_\xi$, and $S_\alpha$, respectively. Compared to the second-best Hyper-HRNet, S3GNet achieves a small but stable performance improvement in several key metrics, including a 0.5\% reduction in MAE (the lower the better), a 1.3\% improvement in PRE, and a 2.1\% improvement in $E_\xi$. These results fully demonstrate the stable performance and strong competitiveness of our S3GNet on the HS-SOD dataset, further proving its potential application in HSOD tasks.

\subsubsection{Qualitative Results}
Fig.~\ref{fig_hssod} shows the qualitative comparison results on HS-SOD dataset. It can be observed that multiple existing methods are prone to problems such as blurred edges, incomplete detection of salient objects, or misdetection, especially in the large object scene in the fourth row. Although S3GNet has not yet achieved complete detection in this scene, it still demonstrates stronger foreground and background differentiation ability by virtue of the joint utilization of spectral and spatial features, and has higher completeness and clarity in the overall detection effect, further reflecting its robustness and effectiveness.

\subsection{Ablation Studies}
To evaluate the effectiveness of each key module, we performed ablation experiments on HSOD-BIT-V2 dataset.

\begin{table*}[t]
\caption{Quantitative Results on RGB-T SOD datasets\label{tab:table7}}
\label{tab:comparison}
\centering
\begin{tabular*}{\linewidth}{@{\extracolsep{\fill}}@{\hspace{2pt}}l|cccc|cccc|cccc}
\toprule
\multirow{2}{*}{Methods} 
& \multicolumn{4}{c|}{VT821} 
& \multicolumn{4}{c|}{VT1000} 
& \multicolumn{4}{c}{VT5000} \\
\cmidrule(lr){2-5} \cmidrule(lr){6-9} \cmidrule(lr){10-13}
& $F_\beta\uparrow$ & $E_\xi\uparrow$ & $S_\alpha\uparrow$ & $\mathrm{MAE} \downarrow$ 
& $F_\beta\uparrow$ & $E_\xi\uparrow$ & $S_\alpha\uparrow$ & $\mathrm{MAE} \downarrow$ 
& $F_\beta\uparrow$ & $E_\xi\uparrow$ & $S_\alpha\uparrow$ & $\mathrm{MAE} \downarrow$ \\
\midrule
MTMR~\cite{MTMR} & 0.662 & 0.815 & 0.725 & 0.108 & 0.715 & 0.836 & 0.706 & 0.119 & 0.595 & 0.795 & 0.680 & 0.114 \\
CPD~\cite{CPD} & 0.710 & 0.837 & 0.827 & 0.057 & 0.834 & 0.902 & 0.906 & 0.032 & 0.741 & 0.867 & 0.848 & 0.050 \\
DCNet~\cite{DCNet} & 0.823 &0.912 & 0.876 & 0.033 & 0.902 & 0.948 & 0.922 & 0.021 & 0.819 & 0.920 & 0.871 & 0.035 \\
CAVER~\cite{pang2023caver} & 0.835 & 0.919 & \textbf{0.891} & 0.033 & 0.909 & 0.945 & \textbf{0.936} & \textbf{0.017} & 0.835 & 0.924 & \textbf{0.892} & 0.032 \\
LSNet~\cite{zhou2023lsnet} & 0.809 &0.911 & 0.878 & 0.033 & 0.887 & 0.935 & 0.925 & 0.023 & 0.806 & 0.915 & 0.877 & 0.037 \\
LAFB~\cite{LAFB} & 0.831 & 0.910 & - & 0.034 & 0.902 & 0.944 & - & 0.019 & 0.847 & 0.925 & - & 0.032 \\
SACNet~\cite{wang2024alignment} & 0.817 & 0.917 & 0.883 & 0.033 & 0.907 & 0.949 & 0.932 & 0.018 & 0.838 & 0.933 & \textbf{0.892} & \textbf{0.030} \\
\midrule
S3GNet (Ours) & \textbf{0.846} & \textbf{0.921} & 0.874 & \textbf{0.032} 
& \textbf{0.923} & \textbf{0.955} & 0.922 & 0.021
& \textbf{0.857} & \textbf{0.938} & 0.876 & 0.034 \\
\bottomrule
\end{tabular*}
\end{table*}

\begin{table}[t]
\caption{Effectiveness Experiments of SAAM\label{tab:table5}}
\centering
\begin{tabular*}{\linewidth}{@{\extracolsep{\fill}}@{\hspace{5pt}}%
>{\centering\arraybackslash}p{1.5cm}@{\hspace{6pt}}
|cccc}
\toprule
Method & $\mathrm{MAE} \downarrow$ & $E_\xi\uparrow$ & $F_\beta\uparrow$ & $S_\alpha\uparrow$ \\
\midrule
Add & 0.025 & 0.828 & 0.619 & 0.761 \\
CBAM & 0.025 & 0.824 & 0.653 & 0.776 \\
WCA & 0.023 & 0.830 & 0.650 & 0.778 \\
CEA & 0.025 & 0.848 & 0.660 & 0.770 \\
SAAM & \textbf{0.021} & \textbf{0.859} & \textbf{0.699} & \textbf{0.801}\\
\bottomrule
\end{tabular*}
\end{table}

\begin{table}[t]
\caption{Effectiveness Experiments of PGRD\label{tab:table6}}
\centering
\begin{tabular*}{\linewidth}{@{\extracolsep{\fill}}@{\hspace{4pt}}%
>{\centering\arraybackslash}p{1cm}
>{\centering\arraybackslash}p{1cm}
>{\centering\arraybackslash}p{1cm}
|cccc}
\toprule
Baseline & GRM & PGRD & $\mathrm{MAE} \downarrow$ & $E_\xi\uparrow$ & $F_\beta\uparrow$ & $S_\alpha\uparrow$ \\
\midrule
\makebox[1cm][c]{\color{green}\ding{51}} & \makebox[1cm][c]{\color{red}\ding{55}} & \makebox[1cm][c]{\color{red}\ding{55}} & 0.045 & 0.780 & 0.533 & 0.701 \\
\makebox[1cm][c]{\color{green}\ding{51}} & \makebox[1cm][c]{\color{green}\ding{51}} & \makebox[1cm][c]{\color{red}\ding{55}} & 0.026 & 0.836 & 0.672 & 0.776 \\
\makebox[1cm][c]{\color{green}\ding{51}} & \makebox[1cm][c]{\color{red}\ding{55}} & \makebox[1cm][c]{\color{green}\ding{51}} & \textbf{0.021} & \textbf{0.859} & \textbf{0.699} & \textbf{0.801} \\
\bottomrule
\end{tabular*}
\end{table}

\begin{figure}[t] 
  \centering
  \includegraphics[width=\linewidth]{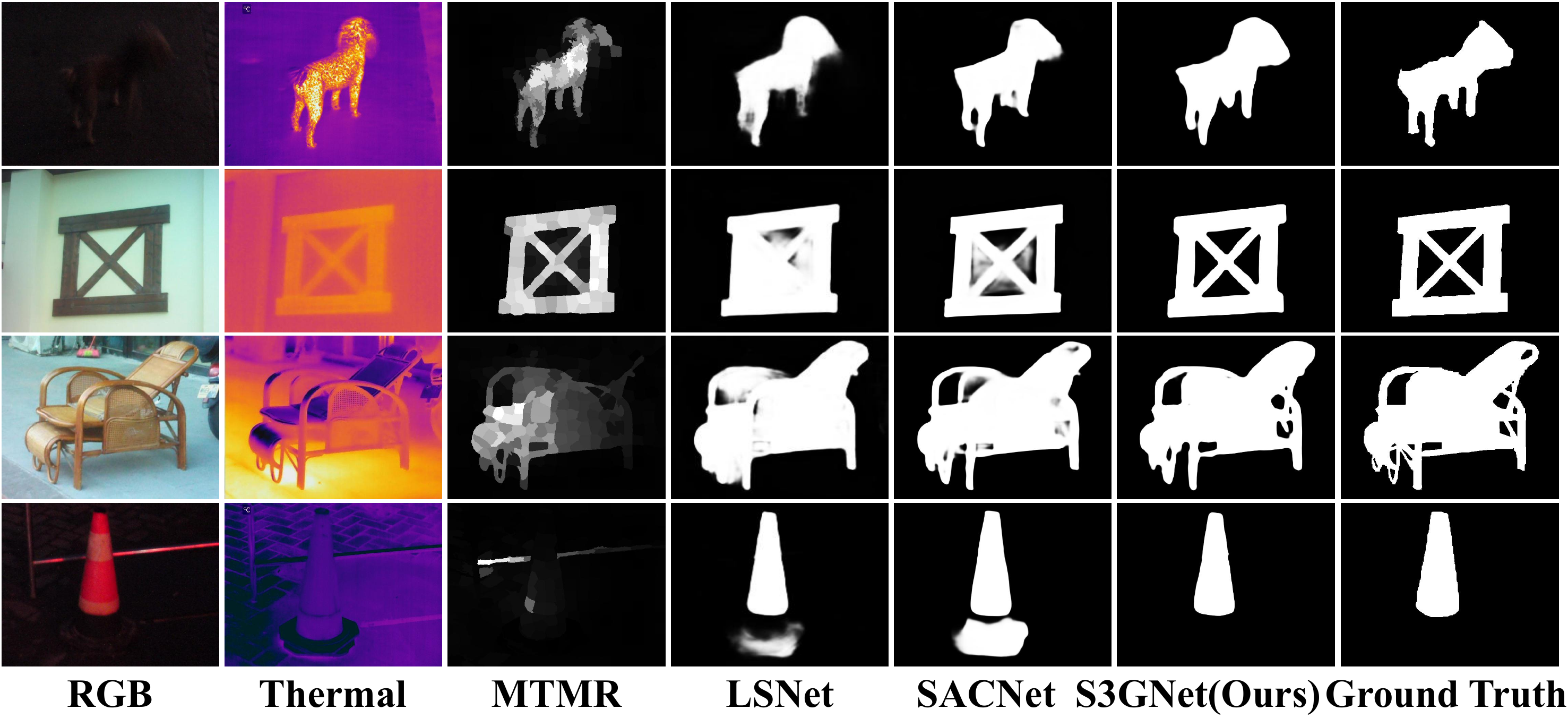} 
  \caption{Qualitative results on RGB-T SOD datasets. S3GNet exhibits excellent significance detection results.}
  \label{fig_rgbt}
\end{figure}

\subsubsection{Effect of Key Components}
To evaluate the contribution of each component in the proposed method, we conducted the experiments shown in Table~\ref{tab:table03}, using only the input spatial information map as a baseline and extending it gradually by introducing different components. The experimental results show that the addition of SSAM, SAAM, and PGRD results in a significant improvement in several metrics of the model, while the best performance is achieved by combining all three. This incremental trend fully validates the effectiveness of each component.

\subsubsection{Effect of Spectral Structure-Aware Module}
To verify the importance of the spectral dimension, we remove the spectral branch and input only pseudo-color images for comparison experiments. As shown in Table~\ref{tab:table3}, the missing spectral information leads to 0.5\%, 4.3\%, 6.5\%, and 3.7\% reduction in MAE, $E_\xi$, $F_\beta$, and $S_\alpha$, respectively, which significantly weakens the detection performance. Fig.~\ref{fig_SSAM} further demonstrates the critical role of SSAM.

\subsubsection{Effect of Different Spectral Derivatives}
To evaluate the effect of different orders of spectral derivatives, we performed comparative experiments as shown in Table~\ref{tab:table13}. The results show that directly using the original spectrum will lead to a decline in detection performance because it is highly sensitive to illumination changes and baseline shifts, making it difficult to provide stable saliency clues. The second-order derivative, although it highlights curvature changes, also amplifies spectral noise and introduces spurious peaks, resulting in limited performance improvement. The mixed derivative improves robustness to some extent, but still suffers from noise amplification. In contrast, the first-order derivative, while suppressing the interference of light, can better capture the inherent spectral change trend of the material, thereby achieving the best detection effect. Therefore, we ultimately choose to use the first-order spectral derivative.

\subsubsection{Effect of Spatial-Spectral Feature Fusion Strategies}
To validate the effectiveness of different spatial and spectral feature fusion strategies, we take the model with the spectral dimension removed as the baseline, and evaluate the introduction of spectral maps or fusion features in only the first layer or in each layer. Table~\ref{tab:table4} shows that direct incorporation of spectral maps instead leads to performance degradation, while the introduction of fusion feature output from SAAM significantly improves detection, especially performing best when fusion features are incorporated in each layer. The analysis concludes that simple splicing is prone to introduce redundancy and lack of semantic alignment, while SAAM realizes spatial-spectral complementarity and multilevel fusion, which improves the model accuracy and robustness. Thus, we finally adopt the introduction of fusion features at each layer as the default strategy.

\subsubsection{Effect of Stream-Aware Attention Module}
To further verify the effectiveness of SAAM, we conducted the experiments shown in Table~\ref{tab:table5} to evaluate direct fusion without attention, using CBAM~\cite{woo2018cbam} to replace SAAM, and using WCA or CEA alone. The results show that the introduction of the attention mechanism is helpful for capturing cross-stream semantic relations and improving detection performance. Compared to a single module, the combination of WCA and CEA has more advantages. WCA focuses on global dependency between streams, while CEA enhances the perception of spatial structure. The two complement and collaborate, enabling SAAM to exhibit stronger robustness and discriminative ability.

\subsubsection{Effect of Progressive Gated Refinement Decoder}

To verify the efficiency of PGRD, we remove the module as a baseline and add GRM and PGRD sequentially. Table~\ref{tab:table6} shows that the addition of GRM improves MAE, $E_\xi$, $F_\beta$, and $S_\alpha$ by 1.9\%, 5.6\%, 13.9\%, and 7.5\%, respectively. On this basis, PGRD introduces a progressive fusion strategy to achieve optimal performance, further validating its crucial role.

\subsection{Extended Experiments on RGB-T SOD}
\subsubsection{Data and Setups}
To further validate the generality and cross-modal adaptation of S3GNet, we extend it to the RGB-T SOD task. The experiments follow the setup of Tu \textit{et al.}~\cite{cong2022does} with the same training set and are evaluated on the VT821, VT1000 and VT5000 test sets. We take $F_\beta$, $E_\xi$, $S_\alpha$ and MAE as the main evaluation metrics, and compare with a variety of advanced methods, systematically verify the robustness and generalization ability of S3GNet in cross-modal saliency detection task.

\subsubsection{Quantitative Results}
As shown in the results of Table~\ref{tab:table7}, S3GNet achieves the highest $F_\beta$ and $E_\xi$ scores on all datasets, significantly outperforming other methods. Although the MAE and $S_\alpha$ on VT1000 and VT5000 are slightly behind, it still maintains leading performance in the RGB-T task without the introduction of SSAM, a key spectral saliency module, verifying the stability and adaptability of S3GNet in cross-modal detection.

\subsubsection{Qualitative Results}
Fig.~\ref{fig_rgbt} shows the qualitative results on RGB-T SOD dataset. Regardless of low illumination or complex background scenes, S3GNet generates remarkable maps with complete targets and clear boundaries, which significantly outperforms other methods and demonstrates its stability and robustness across scenes.

\subsection{Failure Cases and Future Work}
\begin{figure}[t] 
  \centering
  \includegraphics[width=\linewidth]{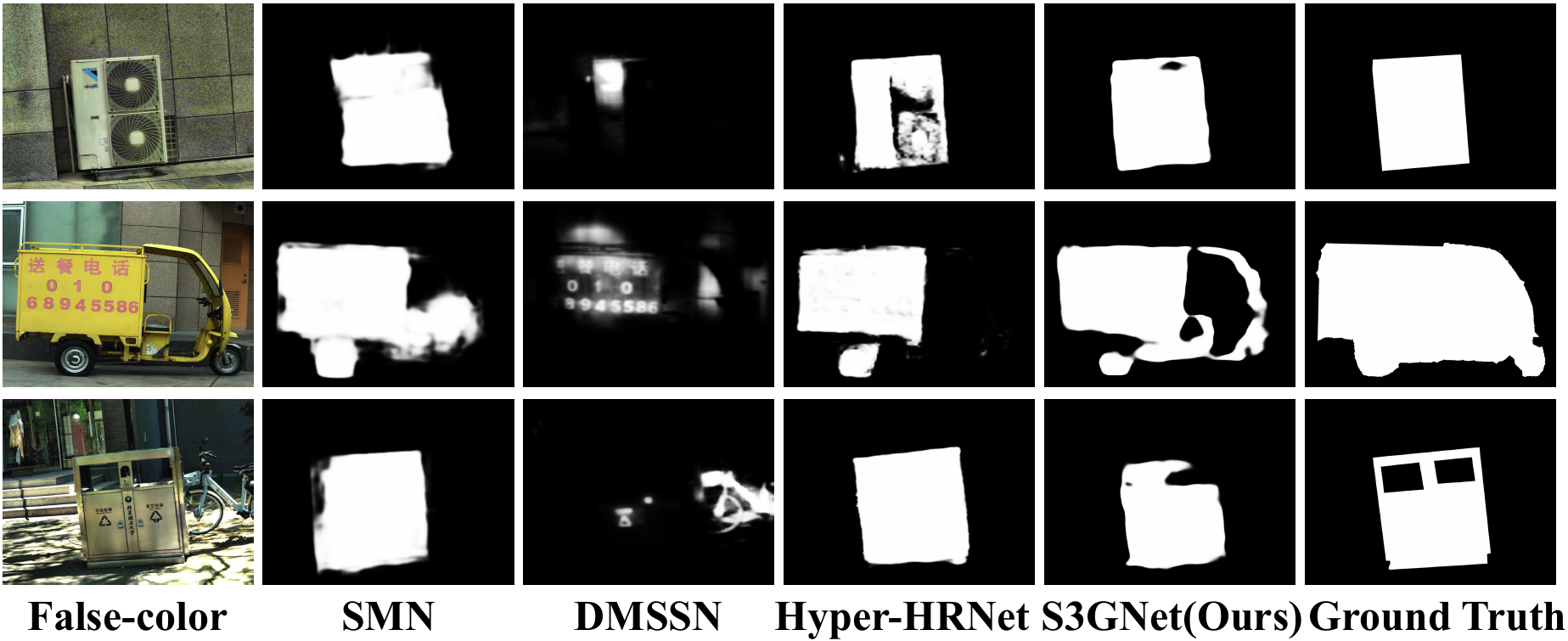} 
  \caption{Visualization results of S3GNet and other advanced methods in some typical failure cases.}
  \label{fig_failure}
\end{figure}

Although S3GNet demonstrates superior performance in the HSOD task by modelling spectral-spatial synergy, it still has limitations in some extremely complex scenarios. As shown in Fig.~\ref{fig_failure}, compared to other methods, our model still suffers from detection incompleteness although it exhibits stronger target integrity. According to our analysis, there are two possible reasons for this limitation: (1) When the salient objects are hollow or have fine structures, the foreground and background regions of the image are highly intertwined, which makes it difficult for existing methods including S3GNet to accurately distinguish and predict them. (2) The training samples for such extreme scenarios are small, making it difficult for the model to learn discriminative features robust enough to cope with the challenge. These failures illustrate that our method is still deficient in fine-grained region understanding.

To this end, we plan to introduce a multi-scale structure-aware module capable of adaptively switching between pixel-level and region-level in subsequent studies. This will allow the model to utilize regional information to resist noise and illumination variations, while also focusing on pixel-level details to inscribe complex boundaries when necessary. We will continue to explore this direction in depth in the future to improve the robustness and generalization of the model in extreme scenarios.

\section{Conclusion}
In this work, we addressed a fundamental challenge in HSOD: the inability of existing models to distinguish essential material properties from incidental spectral noise. Our proposed network, S3GNet, resolves this ambiguity through three synergistic modules. Among them, SSAM builds a robust representation immune to lighting changes. SAAM leverages spatial context to resolve spectral ambiguity, achieving cross-stream complementarity. PGRD preserves critical object details. By tackling the core problem of spectral ambiguity, S3GNet not only sets a new state-of-the-art across multiple benchmarks, but also demonstrates remarkable cross-modal generalization, laying the foundation for more robust saliency detection methods in applications such as remote sensing, precision agriculture and medical imaging.

\bibliographystyle{IEEEtran}

\bibliography{new_ref}




\begin{IEEEbiography}[{\includegraphics[width=1in,height=1.25in,clip,keepaspectratio]{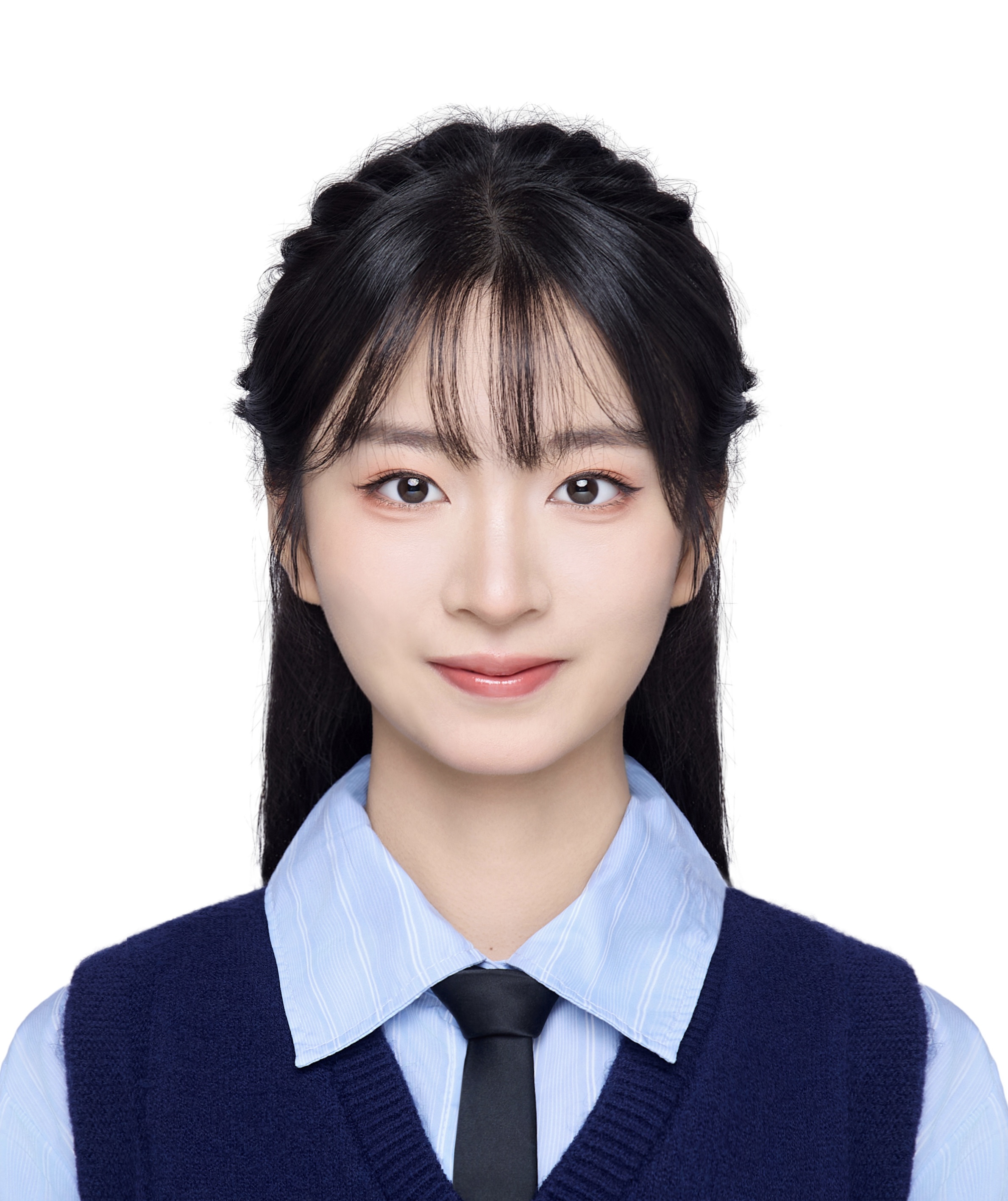}}]{Yanyan Peng} 
received the B.S. degree from the School of Physics, Northeast Normal University, Changchun, Jilin, China, in 2023. She is currently working toward the M.S. degree with the School of Optics and Photonics, Beijing Institute of Technology, Beijing, China. Her research interests include hyperspectral image processing and deep learning.
\end{IEEEbiography}

\begin{IEEEbiography}[{\includegraphics[width=1in,height=1.25in,clip,keepaspectratio]{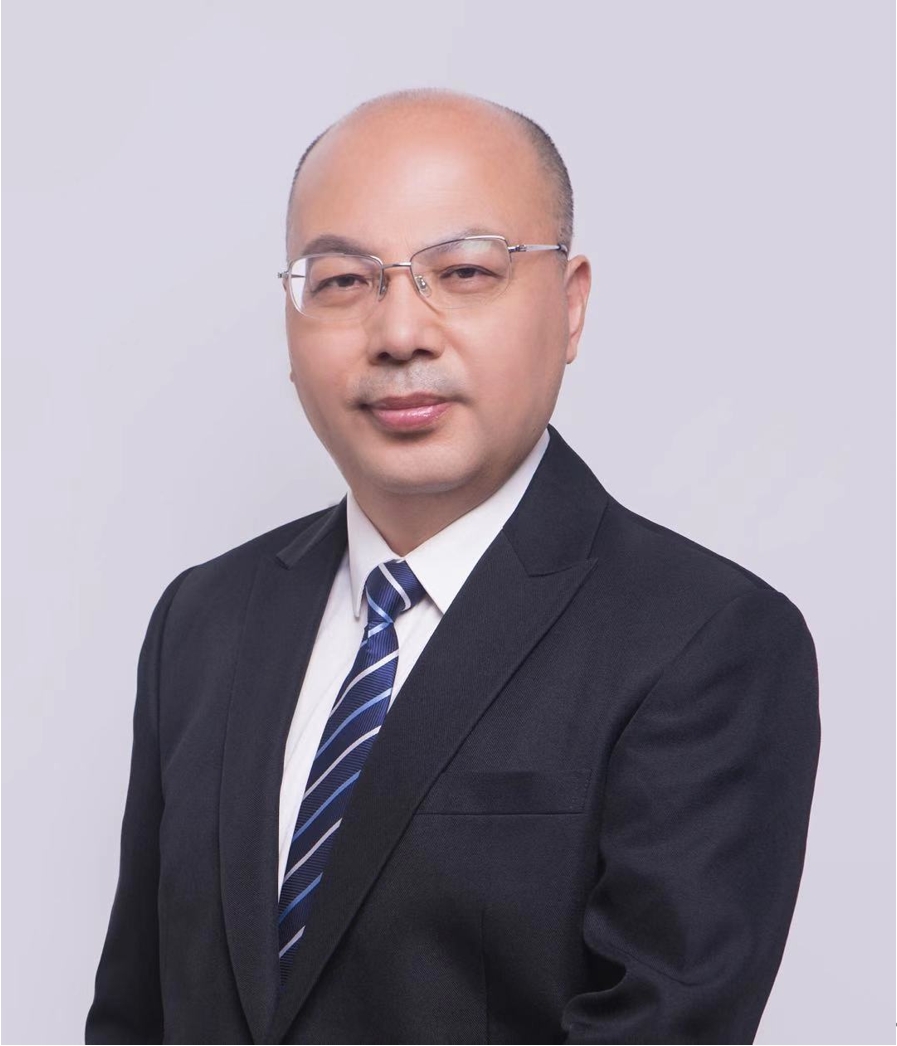}}]{Tingfa Xu} received the Ph.D. degree from the Changchun Institute of Optics, Fine Mechanics and Physics, Changchun, China, in 2004. He is currently a Professor with the School of Optoelectronics, Beijing Institute of Technology, Beijing, China. His research interests include optoelectronic imaging and detection and hyperspectral remote sensing image processing.
\end{IEEEbiography}

\begin{IEEEbiography}[{\includegraphics[width=1in,height=1.25in,clip,keepaspectratio]{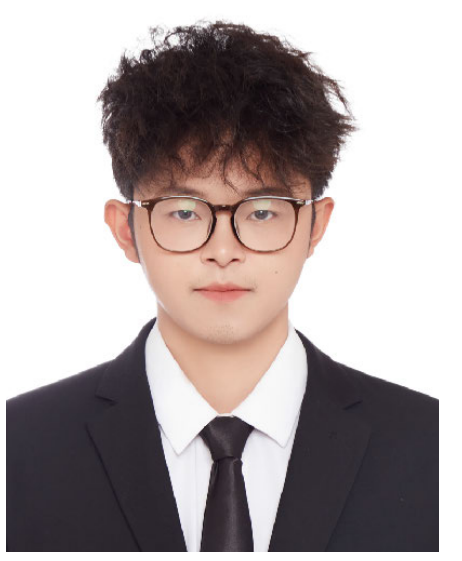}}]{Yao Xiao}
received the M.S. degree in optical engineering with the School of Optics and Photonics, Beijing Institute of Technology, Beijing, China, in 2025. He is currently pursuing the Ph.D. degree in control science and engineering with the College of Control Science and Engineering, Zhejiang University, Hangzhou, China. His research interests include object detection and related computer vision problems.
\end{IEEEbiography}
\vspace{-2em}

\begin{IEEEbiography}[{\includegraphics[width=1in,height=1.25in,clip,keepaspectratio]{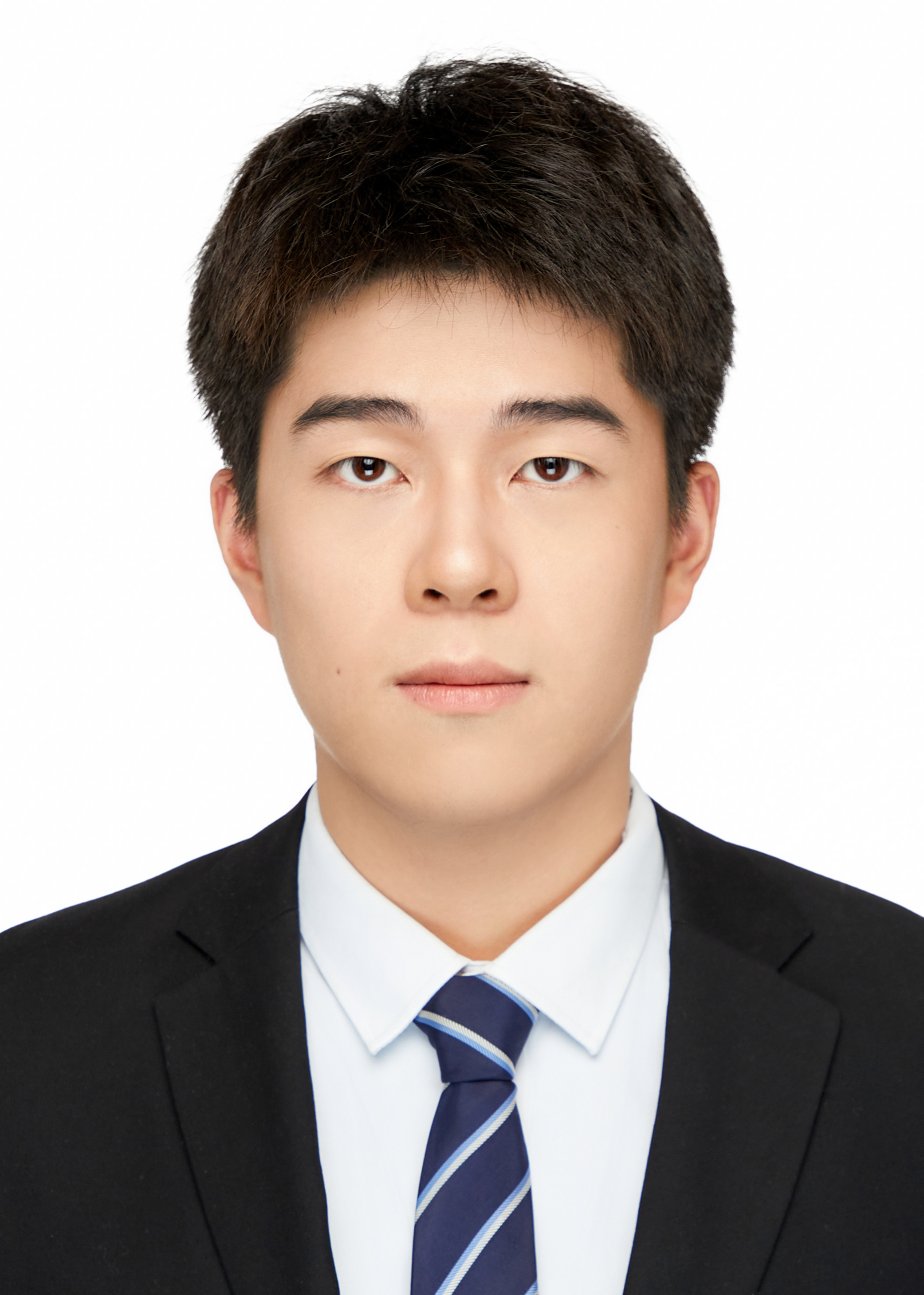}}]{Peifu Liu}
received his B.S. and M.S. degree in engineering from Beijing Institute of Technology in 2021 and 2024, respectively. He is currently pursuing his Ph.D. degree in the School of Optics and Photonics at Beijing Institute of Technology under the supervision of Prof. Xu. His current research interests include hyperspectral image processing and deep learning.
\end{IEEEbiography}
\vspace{-2em}

\begin{IEEEbiography}[{\includegraphics[width=1in,height=1.25in,clip,keepaspectratio]{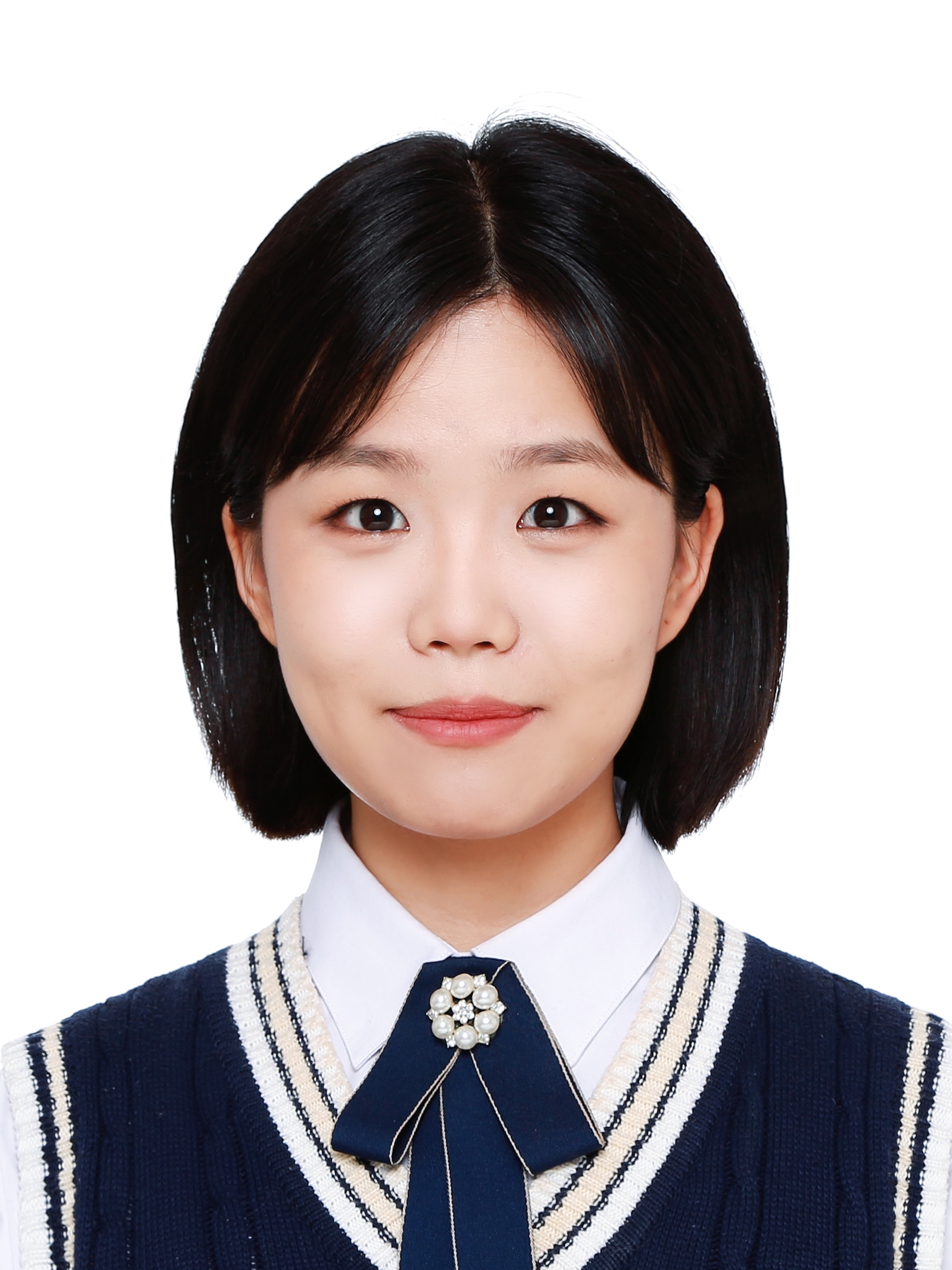}}]{Shuyan Bai}
received the B.S. degree from Beijing Institute of Technology, Beijing, China, in 2023. She is currently working toward the M.S. degree with the School of Optics and Photonics, Beijing Institute of Technology, Beijing, China. Her research interests include hyperspectral camouflaged target detection and deep learning.
\end{IEEEbiography}
\vspace{-2em}

\begin{IEEEbiography}[{\includegraphics[width=1in,height=1.25in,clip,keepaspectratio]{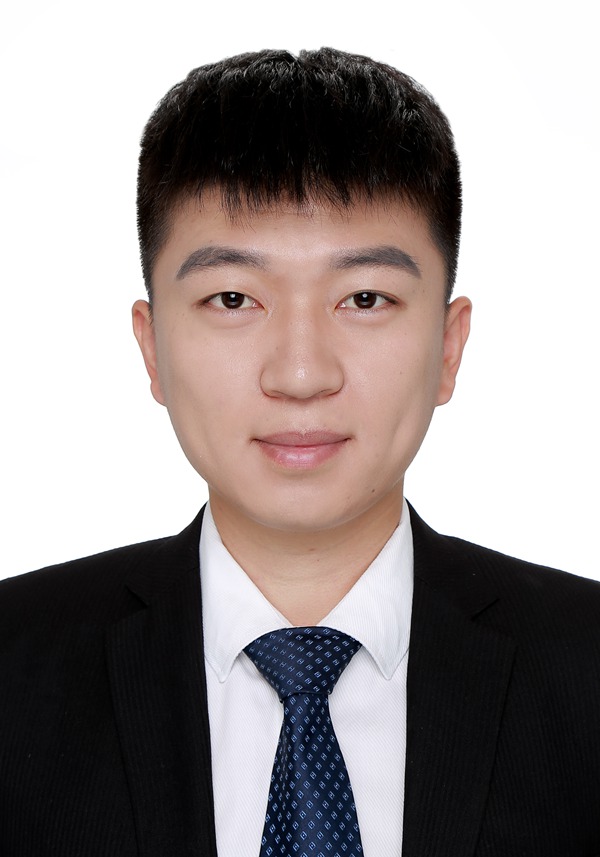}}]{Fengxiang Xu}
is currently pursuing the Ph.D. degree in Electronics and Information at Beijing Institute of Technology. He received the bachelor's degree and master's degree from Beijing Institute of Technology in 2013 and 2017, respectively. His research interests include computer vision, deep learning, and remote sensing image processing.
\end{IEEEbiography}
\vspace{-2em}

\begin{IEEEbiography}[{\includegraphics[width=1in,height=1.25in,clip,keepaspectratio]{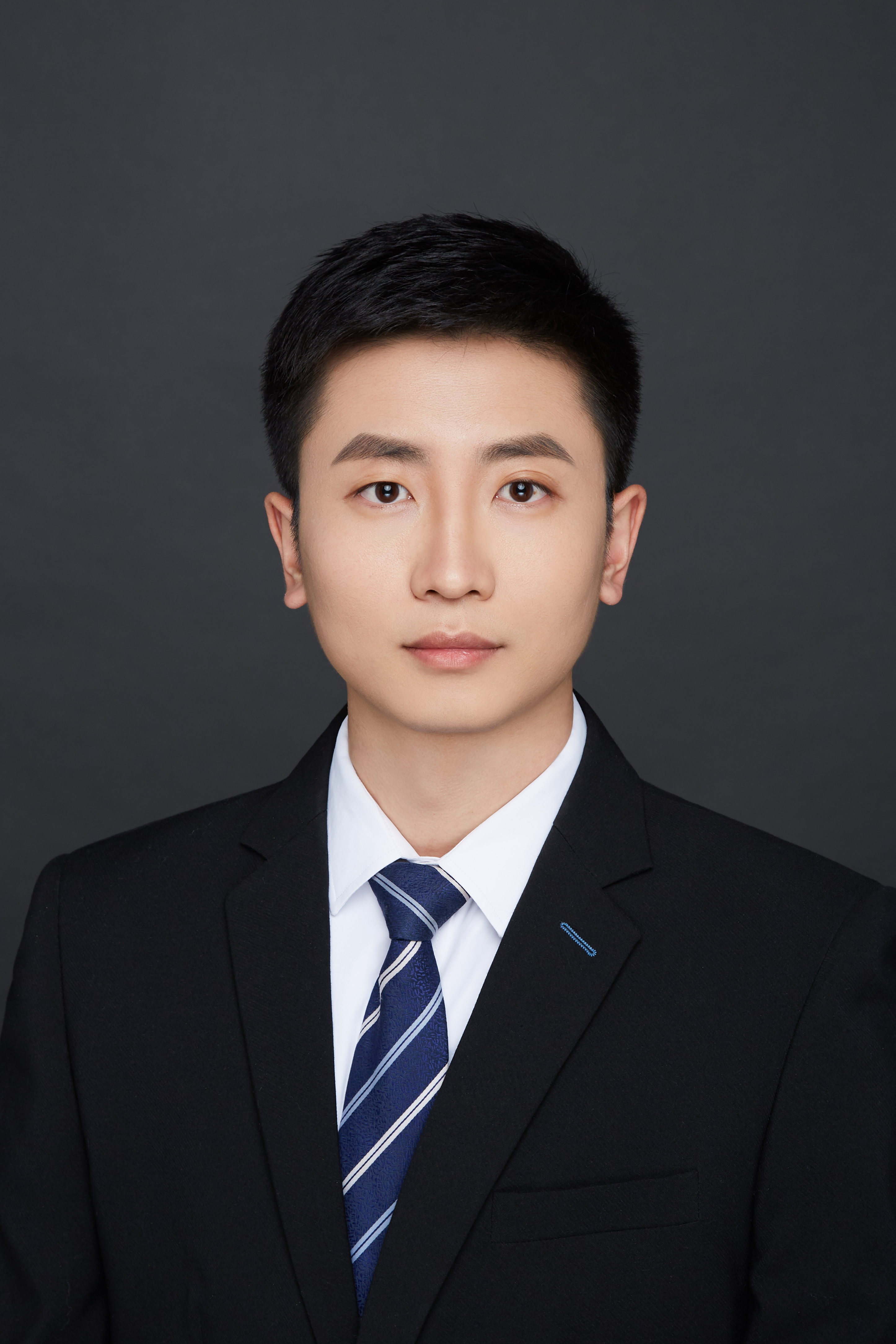}}]{Jianan Li} is currently an associate professor at Beijing Institute of Technology, where he received his B.S. and Ph.D. degree in 2013 and 2019, respectively. From July 2015 to July 2017, he worked as a joint training Ph.D. student at National University of Singapore. From October 2017 to April 2018, he worked as an intern at Adobe Research. His research interests mainly include computer vision and real-time image/video processing.
\end{IEEEbiography}

\end{document}